%% file: main.tex
\documentclass[lettersize,10pt,journal,twoside]{IEEEtran}
\usepackage{ieee_robotics}
\input{config}

\title{\method: Proactive Tactile Control for Robust and Efficient Articulated Object Manipulation}

\author{Zihang Zhao\orcidlink{0000-0003-3215-7152}, Zhenghao Qi\orcidlink{0009-0004-2924-6714}, Yuyang Li\orcidlink{0000-0002-5794-7997}, Leiyao Cui\orcidlink{0009-0009-4925-6983}, Zhi Han\orcidlink{0000-0002-8039-6679}, Lecheng Ruan\orcidlink{0000-0001-5061-3575}, and Yixin Zhu\orcidlink{0000-0001-7024-1545}% <-this % stops a space

\thanks{This work is supported in part by the Brain Science and Brain-like Intelligence Technology--National Science and Technology Major Project (2025ZD0219400), the National Natural Science Foundation of China (62376009), the State Key Lab of General AI at Peking University, the PKU-BingJi Joint Laboratory for Artificial Intelligence, the Wuhan Major Scientific and Technological Special Program (2025060902020304), the Hubei Embodied Intelligence Foundation Model Research and Development Program, and the National Comprehensive Experimental Base for Governance of Intelligent Society, Wuhan East Lake High-Tech Development Zone. (Zihang Zhao and Zhenghao Qi contributed equally to this work. Corresponding authors: Zhi Han, Lecheng Ruan, and Yixin Zhu.)
}%
\thanks{Zihang Zhao, Zhenghao Qi, Yuyang Li, Lecheng Ruan, and Yixin Zhu are with the Peking University, Beijing 100871, China}
\thanks{Zihang Zhao, Zhenghao Qi, Yuyang Li, and Yixin Zhu are also with the State Key Lab of General AI, Peking University, and the Beijing Key Laboratory of Behavior and Mental Health, Peking University, Beijing 100871, China.}%
\thanks{Zihang Zhao is also wth the LeapZenith AI Research, Shanghai 201707, China (email: \texttt{zhaozihang@stu.pku.edu.cn}).}%
\thanks{Zhenghao Qi is also with the Tsinghua University, Beijing 100084, China (emails: \texttt{qi-zh21@mails.tsinghua.edu.cn}).}%
\thanks{Yuyang Li and Yixin Zhu are also with the Embodied Intelligence Lab, PKU-Wuhan Institute for Artificial Intelligence (email: \texttt{y.li@stu.pku.edu.cn}; \texttt{yixin.zhu@pku.edu.cn})}
\thanks{Leiyao Cui is with the University of Chinese Academy of Sciences, Beijing 100049, China, and interned at the Institute for Artificial Intelligence, Peking University, Beijing 100871, China (email: \texttt{cuileiyao24@mails.ucas.ac.cn}).}%
\thanks{Zhi Han is with the University of Chinese Academy of Sciences, Beijing 100049, China (email: \texttt{hanzhi@sia.cn}).}%
\thanks{Data is available online at \texttt{\url{https://tacman-aom.github.io}}.}%
}

\begin{document}

\let\oldtwocolumn\twocolumn
\renewcommand\twocolumn[1][]{%
    \oldtwocolumn[{#1}{
        \vspace{-24pt}
        \centering
        \includegraphics[width=\linewidth]{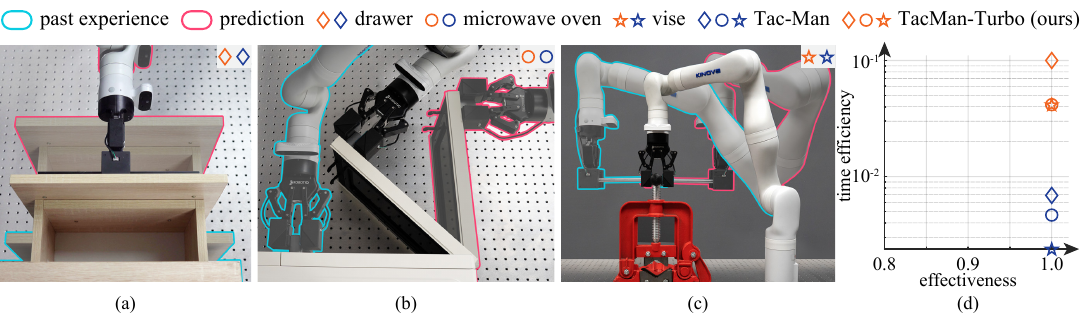}
        \captionof{figure}{\textbf{Robust and efficient tactile-informed manipulation of articulated objects.} \method enables robots to achieve both effective and efficient manipulation of articulated objects through a novel tactile-based control framework. Unlike previous approaches, \method leverages past tactile experience to extract local kinematic information, allowing the controller to predict optimal interaction positions and make proactive adjustments---all without requiring pre-provided object kinematics. This capability significantly enhances manipulation efficiency while maintaining robustness across diverse scenarios. Panels (a-c) demonstrate \method's versatility across various joint mechanisms: (a) manipulating a prismatic-joint drawer in just 10 seconds, (b) operating a revolute-joint microwave oven in 24 seconds, and (c) controlling a complex vise with a helical handle path in 24 seconds. Panel (d) quantifies these improvements, showing that while maintaining \SI{100}{\percent} effectiveness, our method achieves approximately \(15\times\), \(11\times\), and \(18\times\) higher time efficiency than the recent tactile-informed method, \prevmethod~\cite{zhao2024tac}, as measured by average manipulation speeds. These order-of-magnitude improvements in efficiency without compromising reliability mitigate the longstanding trade-off between effectiveness and efficiency in articulated object manipulation.}
        \label{fig:teaser}
    }]
}
\markboth{IEEE Transactions on Automation Science and Engineering}{Proactive Tactile Control for Robust and Efficient Articulated Object Manipulation}

\maketitle

\begin{abstract}
% 1. Basic introduction to the field.
Adept manipulation of articulated objects is essential for robots to operate successfully in human environments. Such manipulation requires both effectiveness---reliable operation despite uncertain object structures---and efficiency---swift execution with minimal redundant steps and smooth trajectories. 
% 2. More detailed background.
Existing approaches struggle to achieve both objectives simultaneously: methods relying on predefined kinematic models lack robustness when encountering structural variations, while the tactile-informed approach achieves robust manipulation but sacrifices efficiency through reactive, step-by-step execute-and-recover cycles.
% 3. The problem being addressed.
To address this challenge, this paper introduces \method, a proactive tactile control framework that unifies the cycles into a continuous control loop.
% 4. Main result.
Our key insight is to interpret tactile signals temporally in addition to spatially: instead of treating contact deviations merely as instantaneous error signals requiring immediate compensation, we analyze sequential tactile observations to reveal local kinematic information. This temporal perspective enables our controller to predict future object states and proactively modulate manipulation velocities, eliminating the previous execute-and-recover cycle. In evaluations across 200 diverse simulated articulated objects and real-world experiments, our approach maintains a near-perfect performance while significantly enhancing time efficiency, action efficiency, and trajectory smoothness (all p-values < 0.0001).
% 5. What the result reveals compared to previous knowledge.
% 6. Results in a more general context.
These results demonstrate that tactile feedback serves a dual purpose: providing not only spatial error signals for reactive control but also critical temporal structural information that enables proactive prediction and control.
% 7. Broader perspective.
By enabling robots to manipulate articulated objects both reliably and efficiently, this work advances robot capabilities for seamless operation in dynamic, human-centric environments.
\end{abstract}

\renewcommand{\abstractname}{Note to Practitioners}
\begin{abstract}
Robots operating in human-centric environments need to adeptly manipulate articulated objects (\eg, drawers and doors), despite uncertain or inaccurate kinematic models, while finishing tasks quickly and smoothly without causing damage. Traditional planners that assume known kinematics are brittle in real-world settings, and recent tactile-reactive controllers, though robust, are slow due to repeated execute-and-recover cycles. \method introduces a proactive tactile control framework built on a temporal interpretation of tactile signals: by extracting local kinematic information from sequential tactile observations, the controller predicts short-horizon handle motion and proactively adjusts end-effector velocity to track these predictions, eliminating dedicated recovery phases. In evaluations across \num{200} simulated articulated objects and real-world experiments, the method achieved near-perfect performance while reducing task completion time by nearly an order of magnitude compared to a strong tactile-based baseline, delivering higher action efficiency and smoother joint trajectories. Overall, \method provides a training-free, drop-in upgrade for tactile manipulation of articulated objects, while preserving robustness and delivering faster, smoother, and more efficient operation without prior object kinematic models.
\end{abstract}


\section{Introduction}

\IEEEPARstart{A}{dept} manipulation of articulated objects, from doors and drawers to various household items, constitutes one of the basic capabilities that robots must master to function effectively in human-centric environments~\cite{wang2023rearrange,jeon2024task}. Success in this domain hinges on two critical objectives: \textit{effectiveness}---the reliable manipulation of objects despite uncertain internal structures, and \textit{efficiency}---the execution of tasks with minimal redundant movements and smooth trajectories~\cite{karayiannidis2012open,karayiannidis2016adaptive}.

Current approaches struggle to achieve both objectives simultaneously. Methods that rely on pre-provided kinematic information demonstrate efficiency in controlled scenarios but often fail when encountering discrepancies between assumed and actual object kinematics. Conversely, while the recent tactile-informed approach by Zhao \etal~\cite{zhao2024tac} successfully achieves robust manipulation despite kinematic uncertainty, it sacrifices efficiency---requiring several minutes to manipulate a single object due to its cautious, step-by-step exploration-compensation mechanism. This fundamental trade-off between effectiveness and efficiency remains unresolved in articulated object manipulation, as further detailed in \cref{sec:related_work-aom}.

To address this challenge, we propose \method, a novel proactive control framework that fundamentally reimagines how tactile feedback can guide manipulation. While leveraging Zhao \etal's insight in using robot-object contact deviations as state representations~\cite{zhao2024tac}, \method introduces a temporal perspective: instead of treating contact deviations merely as spatial errors requiring immediate compensation, we analyze sequential tactile observations as time-series data that reveal local kinematic information. This temporal interpretation enables \method to predict future object states and proactively modulate manipulation velocities to minimize predicted future deviations, streamlining the previous execute-and-recover cycle into a continuous control loop. As illustrated in \cref{fig:teaser}, this novel approach preserves the effectiveness of contact-based manipulation while enabling substantially more efficient execution.

We validate \method in both simulation and real-world settings. Our test scenarios systematically cover articulated objects with diverse joint types---from basic prismatic and revolute joints to complex mechanisms involving concurrent translational and rotational movements. Across \(\boldsymbol{200}\) simulated articulated objects and multiple real-world experiments, \method achieves the same perfect manipulation success rate as the previous tactile-informed approach while significantly improving all efficiency metrics: time efficiency, action efficiency, and trajectory smoothness (all comparisons yielding p-values < \(\boldsymbol{0.0001}\)). These results confirm that \method successfully mitigates the long-standing effectiveness-efficiency trade-off in articulated object manipulation.

Our main contributions are summarized below:
\begin{itemize}[leftmargin=*,noitemsep,nolistsep]
\item \textbf{A temporal interpretation of tactile signals for local kinematic inference:} Rather than treating contact deviations merely as spatial errors requiring immediate compensation, we introduce a temporal perspective that analyzes sequential tactile observations as time-series data. This approach extracts local kinematic information---revealing object motion patterns from contact history---and enables prediction of future object states without requiring pre-provided kinematic models.
\item \textbf{A proactive tactile control framework that unifies manipulation into a continuous process:} Building on our predictive capability, we reformulate articulated object manipulation from a reactive pose-correction problem (minimizing current contact deviations) to a proactive velocity-modulation problem (minimizing predicted future deviations). This reformulation streamlines the execute-and-recover cycle into a unified process, significantly enhancing the manipulation efficiency without compromising the effectiveness.
\end{itemize}

The remainder of this paper is organized as follows: \cref{sec:related_work} positions our work within the broader context of robotic manipulation research. \cref{sec:methods} presents the theoretical framework and implementation details of \method. \cref{sec:simulation,sec:real-world_exp} evaluate \method through simulation studies and real-world experiments, respectively. \cref{sec:discussion} analyzes \method's performance characteristics and discusses future research directions. Finally, \cref{sec:conclusion} summarizes our findings and contributions.

\section{Related Work}\label{sec:related_work}

This section contextualizes \method's contributions within the broader landscape of robotic research. We begin by assessing current approaches to articulated object manipulation, establishing the fundamental trade-off between effectiveness and efficiency that motivates our work (\cref{sec:related_work-aom}). In \cref{sec:related_work-contact_modeling}, we explore contact modeling techniques, highlighting how the proposed method builds upon and advances previous work to address the effectiveness-efficiency trade-off. Next, \cref{sec:related_work-contact_sensing} examines contact sensing technologies, categorizing them into force-based and geometry-based approaches while explaining their compatibility with our method. Finally, \cref{sec:related_work-contact_control} reviews contact control strategies, tracing their evolution from early reactive feedback systems to our proposed proactive tactile control frameworks.

\subsection{Articulated Object Manipulation}\label{sec:related_work-aom}

Approaches to articulated object manipulation can be categorized based on their strategies for addressing the fundamental challenges of effectiveness and efficiency. Traditional methods rely on predefined kinematic models provided manually~\cite{chitta2010planning,burget2013whole}. While these approaches enable efficient manipulation through precise motion planning in controlled environments, they significantly restrict system autonomy by requiring accurate prior knowledge of object kinematics---limiting their effectiveness in real-world scenarios where such information is often unavailable or inaccurate.

To overcome this dependency on manually provided information, subsequent research has focused on the autonomous acquisition of object kinematics. A prominent line of work leverages visual perception to infer kinematic models through geometric reasoning and motion analysis~\cite{hu2017learning,li2020category,zeng2021visual,mittal2022articulated,daiacdc2024,liu2025building,ni2025decompositional}. However, visual information alone often proves insufficient, as visually similar objects may possess fundamentally different internal mechanisms~\cite{zhu2020dark,zeng2021visual,zhao2024tac}. This inherent limitation has led researchers to integrate additional sensing modalities to resolve ambiguity. By combining multi-frame analysis with wrist-mounted force/torque sensors, more accurate parameter estimation of kinematic models becomes possible through active interaction~\cite{karayiannidis2012open,hausman2015active,karayiannidis2016adaptive,martin2022coupled,lv2022sagci,eisner2022flowbot3d,yu2024gamma,wang2024rpmart}. Nevertheless, these methods remain constrained by their underlying assumption that objects conform to predefined joint types (\eg, revolute, prismatic), limiting their effectiveness when encountering complex or unconventional mechanisms.

The persistent challenges in model-based approaches have driven the development of learning-based techniques for articulated object manipulation. Imitation learning enables robots to acquire implicit manipulation strategies from human demonstrations, typically facilitated through teleoperation systems~\cite{pastor2009learning,welschehold2017learning,xiong2021learning,gong2023arnold,zhang2018deep,wong2022error,qin2023anyteleop,fu2024mobile}. This approach naturally captures human manipulation expertise but faces significant challenges in collecting sufficiently diverse and high-quality demonstration datasets~\cite{zheng2022imitation}. While transfer learning techniques have improved generalization to similar objects~\cite{qin2022one,tekden2023grasp}, extending these methods to handle arbitrary articulated objects remains challenging due to the vast diversity of possible kinematic structures.

Reinforcement learning offers an alternative paradigm that attempts to overcome the data collection bottleneck by developing manipulation strategies through repeated interactions in simulated environments~\cite{urakami2019doorgym,xu2022universal,chen2022towards,geng2023partmanip,fan2024reinforcement,honerkamp2025whole}. This approach has been significantly advanced by the creation of large-scale datasets of articulated objects~\cite{mo2019partnet,xiang2020sapien,liu2022akb,geng2023gapartnet,cui2025gapartmanip}, enabling training across diverse object categories. However, the inevitable sim-to-real gap often leads to degraded performance when these methods are deployed in real-world settings, particularly when encountering previously unseen object variations.

The limitations in both model-based and learning-based approaches recently led to the development of tactile-informed manipulation. Zhao \etal~\cite{zhao2024tac} introduced an approach that achieves reliable manipulation of articulated objects by compensating for robot-object contact deviations without requiring kinematic priors. While this method demonstrates robust effectiveness with unknown objects, its reactive execute-and-recover cycle results in significant time overhead. \method advances this foundation through a proactive control framework that unifies the execute-and-recover cycle into a continuous control loop, significantly enhancing manipulation efficiency.

\subsection{Contact Modeling}\label{sec:related_work-contact_modeling}

Contact modeling has emerged as a fundamental approach in robotics research, providing essential information about robot-environment interactions~\cite{siciliano2008springer,li2025taccel,lee2025enhancing,park2025modeling,lee2026improving}. The effectiveness of contact-based strategies has been particularly demonstrated in robotic grasping, where accurate contact models enable reliable manipulation across diverse object geometries~\cite{brahmbhatt2019contactdb,li2024grasp}.

Despite its proven success in grasping, contact modeling remains insufficiently explored for articulated object manipulation---a domain with inherently more complex interaction dynamics. Most existing approaches employ simplified contact assumptions, typically modeling fixed contact points between the robot and the object~\cite{chitta2010planning,burget2013whole,mittal2022articulated,hausman2015active,moses2020visual,lv2022sagci,martin2022coupled}. While this simplification enables efficient computation, it fundamentally fails to capture the dynamic nature of real-world interactions where contact points frequently shift during manipulation. Some progress has been made by relaxing these constraints to permit specific forms of relative motion between the robot and object~\cite{karayiannidis2012open,karayiannidis2016adaptive}, enabling more effective control command generation. However, these approaches still rely on predefined kinematic models, limiting their effectiveness across diverse object types.

Zhao \etal~\cite{zhao2024tac} demonstrated that modeling contact as geometric deviations enables robust articulated object manipulation without requiring prior kinematic models. \method advances this foundation by reinterpreting how these contact deviations are utilized: rather than treating them merely as spatial error signals requiring immediate compensation, we analyze sequential observations temporally to extract local kinematic information, enabling efficient proactive control.

\subsection{Contact Sensing}\label{sec:related_work-contact_sensing}

The successful implementation of \method relies on accurate sensing of contacts between robots and objects, leveraging decades of advances in tactile sensing technology~\cite{yousef2011tactile,luo2025tactile}. Based on their primary output format, tactile sensors can be categorized into two main types: force-based and geometry-based sensors, each offering distinct advantages for manipulation tasks.

Force-based sensors output contact forces by converting mechanical interactions into electrical signals through various transduction mechanisms, including conductive fluid~\cite{fishel2012sensing}, resistance~\cite{ma2014knitted,yang2021non,yu2022all,lee2022predicting}, capacitance~\cite{boutry2018hierarchically,dawood2023learning,zhao2023skin}, and others~\cite{liu2016visual,guo2021visual}. Geometry-based sensors, conversely, directly capture the geometric characteristics of contact interfaces, with vision-based tactile sensors being prominent examples~\cite{yuan2017gelsight,ward2018tactip,li20233,li2024minitac,zhao2025embedding,zhang2025gelstereo,athar2025vibtac,prince2025tacscope}.

Since \method's core functionality centers on contact geometry modeling for local kinematic information extraction, it is naturally compatible with geometry-based sensors that directly provide the required information. Although force-based sensors can also be effectively utilized, they require an additional calibration step using Hooke's law~\cite{gould1994introduction} to convert force measurements into geometric information~\cite{zhao2024tac}. To demonstrate \method's capabilities, in this paper, we implement our system using the GelSight sensor~\cite{yuan2017gelsight}, a well-established geometry-based tactile sensing technology that provides sufficient geometric information about contact surfaces.

\subsection{Contact Control}\label{sec:related_work-contact_control}

Contact control, also known as tactile control, has evolved into a powerful strategy for robotic manipulation by modulating robot motion based on real-time tactile feedback. This approach was pioneered by Weiss \etal~\cite{weiss1987dynamic} and subsequently advanced by Berger and Khosla~\cite{berger1991using}, who provided the first practical implementation demonstrating its potential for precise interaction tasks.

The fundamental challenge in tactile control lies in designing algorithms capable of effectively transitioning from current tactile observations to desired reference states. To address this challenge, researchers have developed two primary approaches: feedback control and direct mapping. Feedback control employs various controllers (\eg, PID controllers) to systematically minimize the difference between observed and desired tactile features, requiring careful selection of task-specific control parameters~\cite{li2013control,kappassov2020touch,lepora2021pose,she2021cable,lloyd2024pose}. Direct mapping, on the contrary, transforms tactile feature differences directly into robot motion commands without explicit parameter tuning. This approach was first introduced by Chen \etal~\cite{chen1995edge}, who proposed the concept of an inverse tactile Jacobian for 2D edge tracking using manually designed features. This foundational framework was subsequently refined and extended by numerous researchers, including Zhang \etal~\cite{zhang2000control} and Sikka \etal~\cite{sikka2005tactile}, who demonstrated its versatility in different manipulation scenarios. More recently, Wilson \etal~\cite{wilson2023cable} advanced the field by introducing motion primitives for cable rearrangement tasks, providing a structured approach to handle deformable objects. Zhao \etal~\cite{zhao2024tac} made a contribution to this paradigm through their implementation of contact point registration, demonstrating robust manipulation of articulated objects despite their kinematic complexity.

The latest evolution in tactile control has seen the emergence of predictive frameworks that anticipate future tactile states rather than reacting only to current observations. For example, Xu \etal~\cite{xu2024letac} employed learning-based techniques to predict future tactile states for reactive grasping. However, such data-driven approaches often face challenges in generalizing across diverse and dynamic tasks due to their dependence on training data distribution. In contrast, \method introduces a model-free, training-free predictive framework grounded in analytical interpretation of tactile sequences. Rather than learning black-box prediction models from data, we extract local kinematic information directly from sequential tactile observations through geometric analysis. This analytical approach enables \method to predict future object states and make proactive adjustments without prior training.

\section{The \acl{ptc}}\label{sec:methods}

At its core, \method is a novel proactive tactile control framework designed for both effective and efficient manipulation of articulated objects. Unlike \textbf{reactive} approaches that respond only to current sensory information, \method extracts local object kinematic patterns from past contact data to predict optimal subsequent interaction positions. This predictive capability enables proactive adjustments that systematically mitigate contact deviations before they significantly impact manipulation performance.

The development of this framework unfolds through four interconnected components. First, \cref{sec:method-notation} establishes the essential mathematical notation and preliminaries that provide the foundation for our approach. Building on this mathematical framework, \cref{sec:method-formulation} formally defines the articulated object manipulation problem, specifically addressing the dual optimization objectives of effectiveness and efficiency. Next, \cref{sec:method-pose_prediction} introduces our novel prediction framework that analyzes sequential patterns in contact deviations to forecast articulated object states during manipulation. Finally, \cref{sec:method-velocity_computation} details how these kinematic predictions are translated into optimal gripper velocity adjustments while adhering to the robot's physical constraints and ensuring consistent contact maintenance.

\begin{figure*}[t!]
    \centering
    \includegraphics[width=\linewidth]{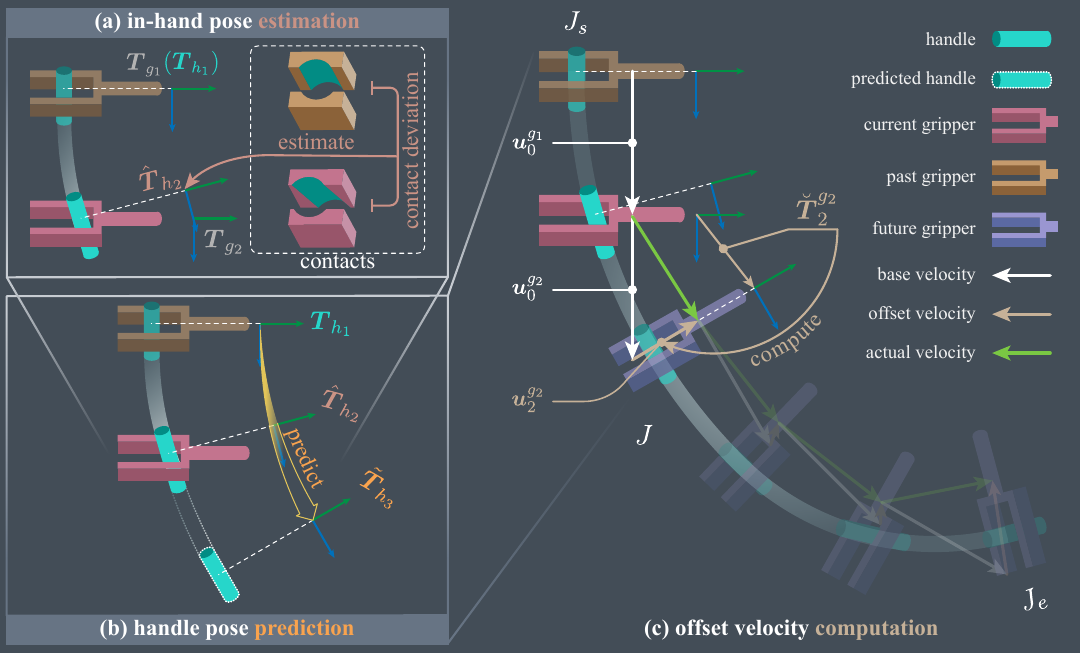}
    \caption{\textbf{Schematic overview of proactive tactile proactive control framework, \method.} Our framework integrates three sequential components that enable proactive manipulation: (a) \textbf{In-hand pose estimation} extracts tactile contact patterns from gripper sensors to determine the relative transformation between current gripper pose \(T_{g_i}\) and handle pose \(T_{h_i}\). (b) \textbf{Handle pose prediction} analyzes sequential pose estimates (\(T_{h_1}\), \(T_{h_2}\)) to extract local kinematic patterns, enabling prediction of the future handle pose \(\tilde{T}_{h_3}\). (c) \textbf{Offset velocity computation} generates the optimal control signal by calculating an offset velocity \(\boldsymbol{u}_i^{g_i}\) that, when combined with base velocity \(\boldsymbol{u}_0^{g_i}\), creates a resultant motion \(\boldsymbol{u}_0^{g_i}+\boldsymbol{u}_i^{g_i}\) that aligns future gripper position with predicted handle motion along path \(J\). By proactively adjusting for potential contact deviations, this approach ensures smooth, continuous manipulation from start point \(J_s\) to end point \(J_e\) without requiring separate correction phases.}
    \label{fig:method}
\end{figure*}

\subsection{Notation}\label{sec:method-notation}

To establish a rigorous foundation for our formulation of effective and efficient manipulation, we define the following mathematical notation:
\begin{itemize}[leftmargin=*,noitemsep,nolistsep]
    \item \(\mathbb{R}\) represents the set of real numbers, with \(\mathbb{R}^{+}\) denoting nonnegative reals. Matrix and vector spaces are denoted as \(\mathbb{R}^{m \times n}\) and \(\mathbb{R}^{n}\) respectively. The special orthogonal group \(SO(n)\) consists of rotation matrices with determinant \(1\), while the special Euclidean group \(SE(n)\) represents rigid body transformations. Additional sets use calligraphic letters (\eg, \(\mathcal{C}\)), with \(\lvert \cdot \rvert\) denoting set cardinality.
    \item Vectors are represented by bold lowercase letters (\eg, \(\boldsymbol{q}\in\mathbb{R}^{n}\)), with subscript \(i\) indicating the \(i\)-th element (\eg, \(q_i\)). The velocity vector \(\boldsymbol{u}\in\mathbb{R}^6\) specifically combines linear velocities \(\boldsymbol{v} = [v_x, v_y, v_z]^{\transpose}\in\mathbb{R}^3\) and angular velocities \(\boldsymbol{\omega} = [\omega_x, \omega_y, \omega_z]^{\transpose}\in\mathbb{R}^3\), where \({\cdot}^{\transpose}\) denotes transpose.
    \item Matrices use bold uppercase letters (\eg, \(\boldsymbol{M}\in\mathbb{R}^{m\times n}\)), with \(m_{ij}\) denoting the entry at row \(i\) and column \(j\). The trace operator is written as \(\text{tr}(\boldsymbol{M})\), and \(\boldsymbol{I}_{n\times n}\) represents the \(n\)-dimensional identity matrix.
    \item For relative pose representation between frames \(\{i\}\) and \(\{j\}\), we use position vector \(\boldsymbol{p}_i^j \in \mathbb{R}^3\) and rotation matrix \(\boldsymbol{R}_i^j \in SO(3)\). When frame \(\{j\}\) is the world frame \(\{w\}\), we omit the superscript. These components combine into the homogeneous transformation matrix:
    \begin{equation*}
        \boldsymbol{T}_i^j = \begin{bmatrix}
            \boldsymbol{R}_i^j & \boldsymbol{p}_i^j \\
              \boldsymbol{0}    & 1
        \end{bmatrix}.
    \end{equation*}
\end{itemize}

\subsection{Problem Formulation}\label{sec:method-formulation}

We now formulate the mathematical framework for effective and efficient tactile-informed manipulation of articulated objects. Consider an articulated object with a movable part and its rigidly attached ``handle,'' as illustrated in \cref{fig:method}. The underlying articulation mechanism constrains the handle's motion to a 1-\ac{dof} trajectory, represented as path \(J\) from initial point \(J_s\) to target point \(J_e\). In our problem setting, only the initial position \(J_s\) is initially known, reflecting real-world scenarios where complete kinematic models are unavailable.

At the start of manipulation, a robot gripper at global pose \(\boldsymbol{T}_{g_1}\) grasps the handle at pose \(\boldsymbol{T}_{h_1}\). For mathematical convenience and without loss of generality, we align the initial handle coordinate frame with that of the robot gripper:
\begin{equation}
    \boldsymbol{T}_{h_1} = \boldsymbol{T}_{g_1}.
    \label{eq:init_handle_pose}
\end{equation}

The physical interaction between the gripper and handle is characterized by their contact state, denoted by \(\mathcal{C}_i\). This state is represented geometrically as a set containing three-dimensional position vectors of activated contact points in the gripper's coordinate frame, denoted as \([p_x^{g_i}, p_y^{g_i}, p_z^{g_i}]^{\transpose}\). Contact points become activated when their normal deformation (illustrated along the \(x\)-axis in \cref{fig:method}) exceeds the threshold \(\epsilon\):
\begin{equation}
    \mathcal{C}_i = \left\{ 
    \begin{bmatrix}
        p_x^{g_i} \\
        p_y^{g_i} \\
        p_z^{g_i}
    \end{bmatrix} \; \middle| \; \lvert p_x^{g_i} \rvert \geq \epsilon \right\},
    \label{eq:contact_points}
\end{equation}
where the threshold \(\epsilon\) is adaptively selected in experiments, with implementation details provided in \cref{sec:real-tactile_sensors}.

The initial contact state \(\mathcal{C}_1\) is assumed stable, providing a secure connection without material damage. Generally, any contact state \(\mathcal{C}_i\) is defined through a mapping function \(f_c\) of handle and gripper poses, as the relative pose between these elements fully determines their interaction:
\begin{equation}
    \mathcal{C}_i = f_c (\boldsymbol{T}_{h_i}, \boldsymbol{T}_{g_i}).
    \label{eq:contact-def}
\end{equation}
In practice, this contact state is directly measured by tactile sensors mounted on the gripper.

Manipulation begins with a constant base velocity \(\boldsymbol{u}_0^{g_i}\) in the gripper's coordinate system, aligned with a known rough direction of the manipulation task. The acquisition of this base velocity and its impact on manipulation efficiency are examined in detail in \cref{sec:discussion}.

Without prior knowledge of the articulation mechanism, the trajectories of the gripper and handle inevitably become misaligned during manipulation. This misalignment produces contact deviations that may result in inefficient or even unstable interactions if not addressed. To mitigate this issue, an offset gripper velocity, denoted as \(\boldsymbol{u}_i^{g_i}\), must be introduced to adapt the gripper's motion to the handle's constrained path.

The key challenge lies in determining this offset velocity. Efficiency considerations preclude simply halting motion to correct current contact deviations---instead, corrections must be planned proactively for subsequent steps, requiring anticipation of how the contact state will evolve. Therefore, the offset velocity cannot simply react to current deviations but must proactively minimize future contact deviations while maintaining the base velocity. Mathematically, this can be formulated as the following optimization problem:
\begin{equation}
    \boldsymbol{u}_i^{g_i} = \argmin_{\boldsymbol{u}_i^{g_i}}  \sum_{(\boldsymbol{c}_1,\boldsymbol{c}_{i+1})\in\mathcal{K}_{1,i+1}}\|\boldsymbol{c}_1- \boldsymbol{c}_{i+1}\|_{2}, 
    \label{eq:formulation}
\end{equation}
where \(\mathcal{K}_{1,i+1}\) defines corresponding points between the reference (initial) and subsequent contact states:
\begin{equation}
    \mathcal{K}_{1,i+1} = \{(\boldsymbol{c}_1, \boldsymbol{c}_{i+1}) \mid \boldsymbol{c}_1 \in \mathcal{C}_1, \boldsymbol{c}_{i+1} \in \mathcal{C}_{i+1}\},
    \label{eq:k}
\end{equation}
subject to the velocity limit constraint:
\begin{equation}
    \boldsymbol{u}^m \preccurlyeq \boldsymbol{u}_0^{g_i} + \boldsymbol{u}_i^{g_i} \preccurlyeq \boldsymbol{u}^M,
    \label{eq:constraints}
\end{equation}
where \(\preccurlyeq\) denotes an element-wise "less than or equal to" relationship between vectors, while \(\boldsymbol{u}^m\) and \(\boldsymbol{u}^M\) represent the lower and upper bounds of the velocity that the robot can physically generate.

Given that the contact state depends solely on the poses of the gripper and handle (as established in \cref{eq:contact-def}), maintaining stable contact requires aligning the gripper's motion with the handle's trajectory. Our solution addresses this fundamental challenge in two complementary steps: First, in \cref{sec:method-pose_prediction}, we develop a method to predict the handle's next-step pose \(\boldsymbol{T}_{h_{i+1}}\) based on sequential contact patterns observed during interaction. Second, in \cref{sec:method-velocity_computation}, we compute the control input \(\boldsymbol{u}_i^{g_i}\) that optimally aligns the gripper's motion with this predicted pose while ensuring compliance with the velocity limit constraints.

\subsection{Handle Pose Estimation and Prediction}\label{sec:method-pose_prediction}

To enable proactive control adjustments, our approach leverages past contact data to predict the optimal position for subsequent object interactions. This predictive capability builds upon two complementary components: first, estimating current handle poses from observed contact deviations and known gripper poses; second, leveraging the temporal sequence of these estimated poses to predict future handle positions. We begin by detailing our pose estimation approach.

At each time step, the handle pose \(\boldsymbol{T}_{h_i}\) relates to the gripper pose \(\boldsymbol{T}_{g_i}\) through their relative transformation:
\begin{equation}
    \boldsymbol{T}_{h_i} = \boldsymbol{T}_{g_i}\boldsymbol{T}^{g_i}_{h_i},
\end{equation}
where \(\boldsymbol{T}^{g_i}_{h_i} \in SE(3)\) represents their relative homogeneous transformation. Initially at \(i = 1\), \(\boldsymbol{T}^{g_1}_{h_1} = \boldsymbol{I}_{4\times4}\) following our initial pose definition (\cref{eq:init_handle_pose}).

For subsequent time steps (\(i > 1\)), assuming no slip between the sensor outer surface and the handle, any relative transformation can be observed as deviations between the current contact state \(\mathcal{C}_i\) and the reference state \(\mathcal{C}_1\). This relationship enables us to estimate \(\boldsymbol{T}^{g_i}_{h_i}\) through the following optimization:
\begin{equation}
    {\hat{\boldsymbol{T}}}^{g_i}_{h_i} = \argmin_{{\hat{\boldsymbol{T}}}^{g_i}_{h_i}\in SE(3)} \sum_{(\boldsymbol{c}_1,\boldsymbol{c}_i)\in\mathcal{K}_{1,i}} \left\| {\hat{\boldsymbol{T}}}^{g_i}_{h_i}
    \begin{bmatrix} \boldsymbol{c}_1 \\ 1
    \end{bmatrix} - \begin{bmatrix} \boldsymbol{c}_i \\ 1
    \end{bmatrix} \right\|_{2},
    \label{eq:transformation}
\end{equation}
where 
\begin{equation}
    \mathcal{K}_{1,i} = \{(\boldsymbol{c}_1, \boldsymbol{c}_{i}) \mid \boldsymbol{c}_1 \in \mathcal{C}_1, \boldsymbol{c}_{i} \in \mathcal{C}_{i}\}
\end{equation}
defines the corresponding contact point pairs between the reference and current states.

This optimization problem \eqref{eq:transformation} can be efficiently solved using the Kabsch algorithm~\cite{kabsch1976solution}, provided the contact state \(\mathcal{C}_i\) satisfies two key constraints:

First, it must contain at least three contact points:
\begin{equation}
    \lvert\mathcal{C}_i\rvert \geq 3.
    \label{eq:num_contact_points}
\end{equation}

Second, these points must be non-collinear. Specifically, there must exist points \(\boldsymbol{p}_1, \boldsymbol{p}_2, \boldsymbol{p}_3 \in \mathcal{C}_i\) such that:
\begin{equation}
     (\boldsymbol{p}_2 - \boldsymbol{p}_1) \times (\boldsymbol{p}_3 - \boldsymbol{p}_1) \neq \boldsymbol{0}.
     \label{eq:noncollinear_constraint}
\end{equation}

Given these constraints, we can estimate the handle pose \(\hat{\boldsymbol{T}}_{h_i}\) at each time step \(i\) by combining the gripper pose with the estimated relative transformation:
\begin{equation}
    \hat{\boldsymbol{T}}_{h_i} = \boldsymbol{T}_{g_i} \hat{\boldsymbol{T}}^{g_i}_{h_i}.
\end{equation}

Building upon these estimated poses, we now focus on predicting future handle positions. Under the assumptions of locally smooth object trajectories and sufficiently high sampling frequencies, consecutive handle poses exhibit only slight variations, primarily driven by the object's instantaneous velocity. This local smoothness makes the constant velocity model a physically faithful approximation of the short-horizon dynamics~\cite{davison2007monoslam,mur2015orb,scholler2020constant}. Moreover, compared to higher-order alternatives (\eg, acceleration or jerk models), the constant-velocity formulation requires estimating fewer derivatives from finite-difference observations, which inherently reduces sensitivity to measurement noise. Based on this model, the next handle pose \({\tilde{\boldsymbol{T}}}_{h_{i+1}}\) is predicted as:
\begin{equation}
    {\tilde{\boldsymbol{T}}}_{h_{i+1}} = {\hat{\boldsymbol{T}}}_{h_{i}} \boldsymbol{T}_{u},
\end{equation}
where \(\boldsymbol{T}_{u}\) represents the velocity effects, which are assumed to remain constant over short time intervals. Given consistent sampling intervals, \(\boldsymbol{T}_{u}\) is computed from the two most recent pose estimates:
\begin{equation}
    \boldsymbol{T}_{u} = ({\hat{\boldsymbol{T}}}_{h_{i-1}})^{-1} {\hat{\boldsymbol{T}}}_{h_{i}}.
    \label{eq:t_u}
\end{equation}

While this constant velocity model effectively predicts handle poses, it may introduce minor residual errors relative to ground truth due to the omission of higher-order effects such as acceleration. However, these residual errors are detected by subsequent in-hand pose estimation and corrected during the next prediction step, preventing error accumulation. Moreover, for scenarios involving more dynamic motion or reduced sampling frequencies, this prediction framework can be extended to account for higher-order motion effects by incorporating additional past data---for example, transitioning to a constant acceleration model by using one more observation to account for velocity changes.

As evident from \cref{eq:t_u}, the proposed constant velocity prediction requires at least two previous observations (\(i \geq 2\)). For the initial step (\(i = 1\)), when past data is unavailable, the system relies solely on the base velocity \(\boldsymbol{u}_0^{g_1}\) defined in \cref{sec:method-formulation}. This integration of pose estimation and prediction provides the foundation for computing appropriate gripper velocities, which we address in the following section.

\subsection{Offset Velocity Computation}\label{sec:method-velocity_computation}

Having predicted the next handle pose, we now develop a method for computing the offset gripper velocity that proactively compensates for contact deviations while maintaining operational efficiency. In the ideal scenario, the solution is straightforward: align the gripper's next pose precisely with the predicted handle pose to eliminate any potential contact deviation:
\begin{equation}
    \boldsymbol{T}_{g_{i+1}} = \tilde{\boldsymbol{T}}_{h_{i+1}}.
\end{equation}

This desired alignment requires the gripper to execute a specific transformation over a sampling interval \(\Delta t \in \mathbb{R}_{>0}\). We compute this required transformation relative to the gripper's current coordinate frame:
\begin{equation}
   \breve{\boldsymbol{T}}_{i}^{g_i}  = ({\boldsymbol{T}}_{g_{i}})^{-1} {\boldsymbol{T}}_{g_{i+1}}.
\end{equation}

From this transformation, we derive the required velocity components:
\begin{equation}
    \boldsymbol{u}_i^{g_i} = \frac{\log(\breve{\boldsymbol{T}}_{i}^{g_i})^{\vee}}{\Delta t} - \boldsymbol{u}_0^{g_i},
\end{equation}
where \(\log(\cdot)\) denotes the logarithmic map and the operator \(\cdot^{\vee}\) transforms the skew-symmetric matrix into vector form. 

While the above formulation provides the theoretically optimal velocity, practical implementation must account for the limitations imposed by robot hardware---specifically the maximum gripper velocities determined by the robot's joint motor limits, as defined in \cref{eq:constraints}. The relationship between joint velocities \(\dot{\boldsymbol{q}}_i\in\mathbb{R}^{n}\) and the gripper velocity \(\boldsymbol{u}_i^{g_i}\) is given by:
\begin{equation}
     \boldsymbol{J}(\boldsymbol{q}_i) \dot{\boldsymbol{q}}_i = \boldsymbol{u}_i^{g_i} + \boldsymbol{u}_0^{g_i},
\end{equation}
where \(\boldsymbol{J}(\boldsymbol{q}_i)\in\mathbb{R}^{6\times n}\) is the robot Jacobian matrix. For analytical tractability, we assume \(n \geq 6\) and that \(\boldsymbol{J}(\boldsymbol{q}_i)\) maintains row full rank. Over small time intervals, treating \(\boldsymbol{q}_i\) as constant allows us to solve for joint velocities:
\begin{equation}
    \dot{\boldsymbol{q}}_i = \boldsymbol{J}^{\dagger}(\boldsymbol{q}_i) (\boldsymbol{u}_i^{g_i} + \boldsymbol{u}_0^{g_i}),
\end{equation}
using the pseudoinverse:
\begin{equation}
    \boldsymbol{J}^{\dagger}(\boldsymbol{q}_i) = \boldsymbol{J}(\boldsymbol{q}_i)^{\transpose}(\boldsymbol{J}(\boldsymbol{q}_i)\boldsymbol{J}(\boldsymbol{q}_i)^{\transpose})^{-1}.
\end{equation}

To ensure joint velocities remain within their maximum limits \(\dot{\boldsymbol{q}}_M\), we introduce a scaling factor \(\alpha\):
\begin{equation}
    \alpha = \min\left(\min\left(\left\{\frac{\dot{q}_{M_j}}{\max(\lvert\dot{q}_{i_j}\rvert,\delta)}\beta \; \middle| \; j = 1,\ldots, n\right\}\right),1\right),
\end{equation}
where \(\beta\) is a safety factor and \(\delta\) represents a small positive constant introduced to prevent division by zero. In this study, we select \(\beta = 2\) and \(\delta = 1\times10^{-4}\) across all the experiments. This scaling yields the final executable offset velocity command:
\begin{equation}
    \boldsymbol{u}_i^{g_i} = \alpha\boldsymbol{u}_i^{g_i} + (\alpha-1)\boldsymbol{u}_0^{g_i}
\end{equation}
with the base velocity also scaled by \(\alpha\). Consequently, to achieve the desired handle position \(\tilde{\boldsymbol{T}}_{h_{i+1}}\) while respecting joint velocity limits, the execution duration must be adjusted from \(\Delta t\) to:
\begin{equation}
    \Delta t' = \frac{\Delta t}{\alpha}.
\end{equation}

While this scaling may marginally reduce manipulation efficiency, it represents a necessary adjustment to ensure the robot operates within its physical constraints, particularly when addressing large contact deviations that require substantial compensation.

\begin{figure*}[t!]
    \centering
    \begin{subfigure}[b]{\linewidth}
        \centering
        \includegraphics[width=\linewidth]{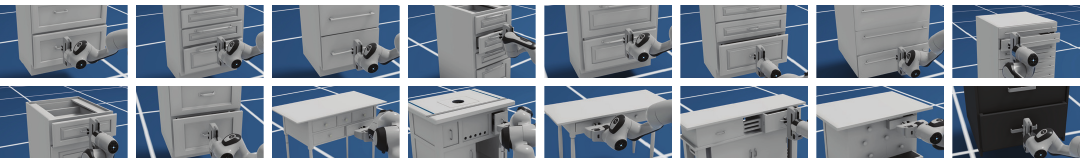}
        \caption{Prismatic-joint objects with linear manipulation paths.}
        \label{fig:simulation-setup-a}
    \end{subfigure}%
    \\%
    \begin{subfigure}[b]{\linewidth}
        \centering
        \includegraphics[width=\linewidth]{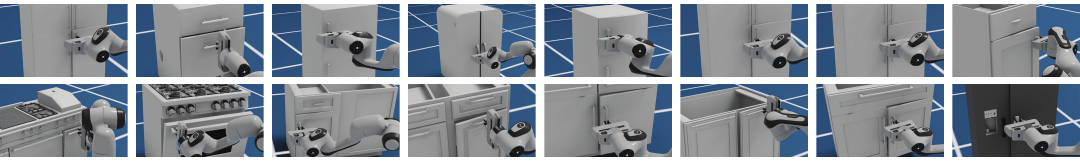}
        \caption{Revolute-joint objects with rotational manipulation paths.}
        \label{fig:simulation-setup-b}
    \end{subfigure}%
    \\%
    \begin{subfigure}[b]{\linewidth}
        \centering
        \includegraphics[width=\linewidth]{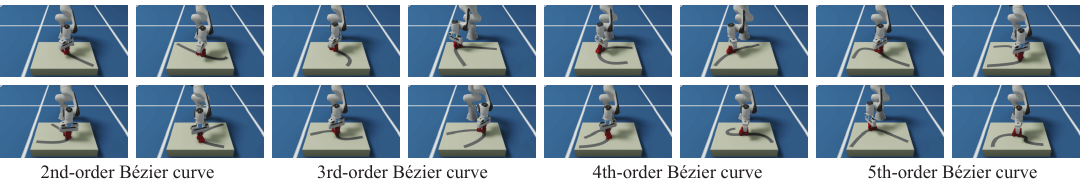}
        \caption{Complex articulation objects with trajectories defined by Bézier curves of different orders.}
        \label{fig:simulation-setup-c}
    \end{subfigure}%
    \caption{\textbf{Simulation test environment and object categories.} Our evaluation encompasses three progressively complex articulated object categories: (a) 50 prismatic-joint objects featuring linear motion paths, representing the most kinematically straightforward mechanisms; (b) 50 revolute-joint objects with circular motion paths, introducing rotational complexity; and (c) 100 complex articulated objects with trajectories derived from Bézier curves of orders 2--5, simulating sophisticated real-world mechanisms with variable curvature. This systematically designed test suite enables rigorous performance assessment across a spectrum of kinematic complexity. Each case is standardized using initial robot configurations, ensuring fair comparison across manipulation strategies.}
    \label{fig:simulation-setup}
\end{figure*}

\section{Simulation Studies}\label{sec:simulation}

To thoroughly evaluate \method, we conducted extensive simulation studies focusing on both manipulation effectiveness and efficiency. This section presents our experimental approach and findings in four parts: we first detail our simulation setup (\cref{sec:simulation-setup}), establish comparative baselines (\cref{sec:simulation-baselines}), define quantitative evaluation metrics (\cref{sec:simulation-metrics}), and analyze the performance results (\cref{sec:simulation-results}).

\subsection{Simulation Setup}\label{sec:simulation-setup}

Our evaluation framework integrates three core components designed to enable comprehensive assessment of manipulation capabilities under controlled yet realistic conditions.

\paragraph*{Simulation environment} 

We implemented our experiments in NVIDIA Isaac Sim, operating at \SI{60}{\hertz} to ensure sufficient temporal resolution for capturing dynamic contact interactions. Critical to our study is the contact simulation method developed by Zhao \etal~\cite{zhao2024tac}, which enables accurate modeling of tactile feedback by monitoring displacement patterns within designated contact regions. This approach captures the nuanced contact dynamics that emerge during the manipulation of articulated objects.

\paragraph*{Test articulated objects} 

To systematically evaluate performance across varying kinematic structures, we assembled a three-category test suite with progressively increasing trajectory complexity:
\begin{itemize}[leftmargin=*,noitemsep,nolistsep]
    \item 50 objects with prismatic joints from PartNet-Mobility dataset~\cite{xiang2020sapien} (\cref{fig:simulation-setup-a}), featuring linear motion paths with varying physical characteristics.
    \item 50 objects with revolute joints also from PartNet-Mobility (\cref{fig:simulation-setup-b}), representing rotational articulations that require continuous adjustment of manipulation direction.
    \item 100 custom-generated objects with Bézier curve trajectories designed to assess performance on complex, non-linear paths. This category includes 25 curves each of orders 2, 3, 4, and 5, sampled on a 2D plane and constrained to avoid self-intersections. Each curve defines a manipulation path on a 3D playboard, with a toy train positioned at the starting point serving as the handle (\cref{fig:simulation-setup-c}).
\end{itemize}

Such a design from simple to complex kinematics provides systematic evaluation across a spectrum of trajectory predictability and curvature characteristics. The generation process for the Bézier-based objects followed protocols in Zhao \etal~\cite{zhao2024tac}, producing diverse trajectory complexities that challenge predictive capabilities in distinct ways.

\paragraph*{Robot} 

Our experiments employ a 7-\ac{dof} Franka robotic arm, selected for its ability to execute full 6-\ac{dof} Cartesian motions---a capability essential for accommodating the diverse movement requirements of our test objects. We also integrate the B* algorithm to place the robot, ensuring that hand trajectories during manipulation remain within the robot workspace~\cite{zhao2025b}. We implement motion control using NVIDIA Isaac Sim's built-in RMPFlow algorithm~\cite{cheng2021rmpflow}. Each experimental trial begins with a standardized initialization: the robot establishes a secure grasp on the target handle, creating a well-defined reference contact state \(\mathcal{C}_1\) that serves as the foundation for subsequent manipulation and deviation measurements.

\subsection{Baseline Selection}\label{sec:simulation-baselines}

For comparative analysis, we adopt \prevmethod~\cite{zhao2024tac} as our primary baseline, a selection driven by its demonstrated superiority in manipulation effectiveness. The key architectural distinction between \prevmethod and our proposed \method framework lies in their fundamental approach to manipulation control. \prevmethod implements a two-stage \textbf{reactive} paradigm: an execution stage that performs the primary manipulation, followed by a separate recovery stage that corrects contact deviations after they occur. In contrast, \method integrates these functions into a unified \textbf{proactive} process that anticipates and preemptively addresses potential deviations, enabling more fluid and continuous manipulation.

Beyond the primary baseline, we evaluate \method against three additional baselines spanning traditional pre-planned methods, impedance control strategies, and learning-based approaches to contextualize our contribution within the broader literature. The pre-planned method leverages known kinematics to generate trajectories. To model parameter estimation uncertainty in real-world deployment, we perturb the ground-truth kinematic parameters with Gaussian noise (variance equals \(0.01\)), a level consistent with the accuracy of current vision-based estimation methods~\cite{yu2024gamma,wang2024rpmart}. The impedance control baseline operates under the same prior-free setting as \prevmethod but excludes tactile feedback, instead relying on impedance adaptation to accommodate deviations. Finally, the learning-based baseline employs FlowBot3D~\cite{eisner2022flowbot3d} in a zero-shot setting, which is trained to use visual feedback for manipulation guidance.

To ensure a fair and rigorous comparison, both methods operate under identical initialization conditions as specified in \cref{sec:simulation-setup}, including consistent robot configuration, initial grasp state, and interaction direction. This standardized experimental protocol effectively isolates the impact of their differing control strategies, ensuring that performance differences directly reflect the fundamental distinction.

\begin{figure*}[t!]
    \centering
    \includegraphics[width=\linewidth]{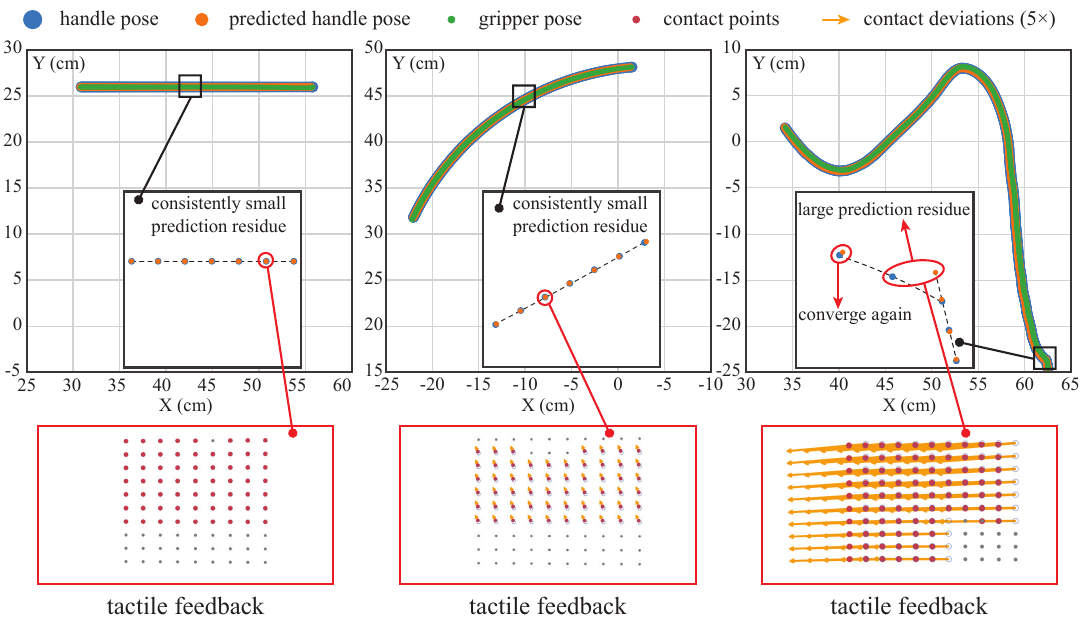}
    \caption{\textbf{Qualitative results of simulation studies.} This visualization demonstrates \method's performance across three representative cases from our object categories: prismatic (left), revolute (middle), and complex articulation (right). The top row displays trajectory tracking performance with actual handle positions (blue), predicted handle positions (orange), gripper positions (green), and contact points (red), while the bottom row shows corresponding tactile feedback patterns. For objects with consistent kinematic patterns (prismatic and revolute), \method's prediction mechanism generates remarkably accurate estimates with minimal residual error, enabling precise trajectory alignment. When confronting complex articulations with sudden directional changes (right column), prediction accuracy temporarily decreases at inflection points, but the system rapidly converges back to the optimal path through iterative correction. The magnified insets highlight these prediction characteristics, demonstrating how \method maintains effective manipulation even when faced with kinematically challenging scenarios. Contact deviation vectors (orange arrows, magnified 5× for visibility) further illustrate how the system continuously adjusts to maintain optimal contact. More qualitative results are available in the \href{\suppUrlSI}{Supplementary Video S1}.}
    \label{fig:sim-results-qualitative}
\end{figure*}

\begin{figure*}[t!]
    \centering
    \begin{subfigure}[b]{\linewidth}
        \centering
        \includegraphics[width=\linewidth]{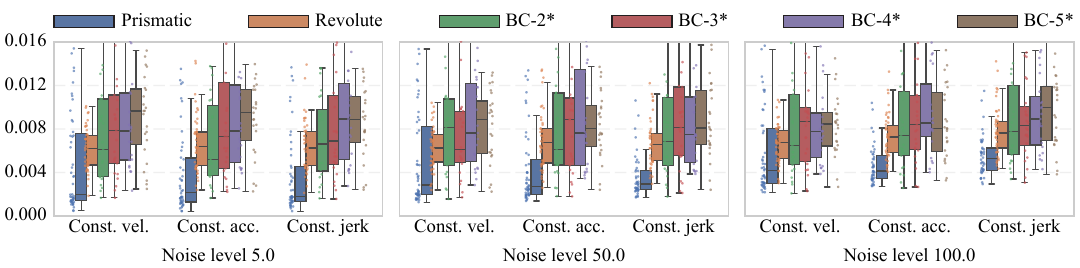}
        \caption{Distribution of mean errors.}
        \label{fig:sim-results-prediction-a}
    \end{subfigure}%
    \\%
    \begin{subfigure}[b]{\linewidth}
        \centering
        \includegraphics[width=\linewidth]{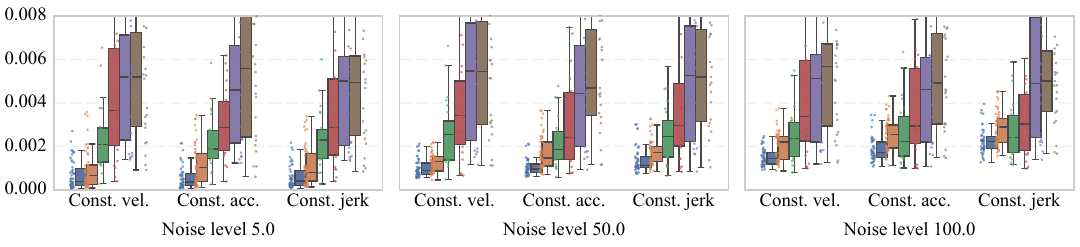}
        \caption{Distribution of error standard deviations.}
        \label{fig:sim-results-prediction-b}
    \end{subfigure}%
    \caption{\textbf{Prediction errors of various predictive models.} The prediction error distributions for three kinematic models---constant velocity, constant acceleration, and constant jerk---are presented and compared. All methods demonstrate comparable performance overall. However, higher-order models exhibit superior accuracy under low-noise conditions, as they better capture the underlying motion dynamics. Conversely, these same models show degraded performance in high-noise environments due to the cumulative effects of numerical differentiation, which amplifies measurement noise with each successive derivative computation. Box plots display time distributions with medians (center lines), interquartile ranges (boxes), and distribution extent (whiskers at \(1.5\times\)IQR). BC-n represents Bézier curves of order n.}
    \label{fig:sim-results-prediction}
\end{figure*}

\begin{table}[t!]
\centering
\small
\color{black}
\setlength{\tabcolsep}{5pt}
\caption{\textbf{Maximum prediction errors in centimeters comparing three prediction methods across articulated object categories.}}
\label{tab:max_value}
\begin{tabular}{@{}lccccccc} 
\toprule
\textbf{Method} & \multicolumn{1}{l}{\begin{tabular}[c]{@{}l@{}}\textbf{\textbf{Noise}}\\\textbf{\textbf{level}}\end{tabular}} & \textbf{Pri.} & \textbf{Rev.} & \textbf{BC-2} & \textbf{BC-3} & \textbf{BC-4} & \textbf{BC-2} \\ 
\midrule
\textbf{Const. vel.} & \multirow{3}{*}{5.0} & 1.9 & 1.6 & 2.2 & 3.5 & 3.9 & 3.3 \\
\textbf{Const. acc.} &  & 2.2 & 1.6 & 3.1 & 3.9 & 4.3 & 4.5 \\
\textbf{Const. jerk} &  & 2.2 & 1.8 & 2.2 & 4.0 & 4.2 & 3.9 \\ 
\midrule
\textbf{Const. vel.} & \multirow{3}{*}{50.0} & 1.7 & 1.8 & 2.5 & 3.7 & 4.2 & 3.2 \\
\textbf{Const. acc.} &  & 1.8 & 2.2 & 1.9 & 3.7 & 3.6 & 3.2 \\
\textbf{Const. jerk} &  & 2.1 & 2.1 & 1.9 & 3.2 & 4.0 & 4.1 \\ 
\midrule
\textbf{Const. vel.} & \multirow{3}{*}{100.0} & 2.1 & 2.2 & 2.5 & 3.4 & 4.0 & 3.4 \\
\textbf{Const. acc.} &  & 2.0 & 2.5 & 2.6 & 2.5 & 3.1 & 3.4 \\
\textbf{Const. jerk} &  & 2.1 & 2.5 & 2.1 & 2.9 & 3.2 & 2.9 \\
\bottomrule
\end{tabular}
\end{table}

\begin{table}[b!]
    \centering
    \small
    \color{black}
    \setlength{\tabcolsep}{3.8pt}
    \caption{\textbf{Success rate (\%) comparing \method and baseline methods across articulated object categories.} Results are reported as mean (standard deviation).}
    \label{tab:sim-results-success_rate}
    \begin{tabular}{@{}l*{6}{S[table-format=3.0\pm2.0, retain-zero-uncertainty=true]}@{}}
        \toprule
        \textbf{Method} & \textbf{Pri.} & \textbf{Rev.} & \textbf{BC-2} & \textbf{BC-3} & \textbf{BC-4} & \textbf{BC-5}  \\
        \midrule
        \textbf{Planing} & 96\pm16 & 93\pm21 & 71\pm29 & 39\pm37  & 53\pm40 & 43\pm36 \\
        \textbf{Imped.} & 81\pm33 & 48\pm28 & 32\pm28 & 32\pm22  & 20\pm23 & 43\pm27 \\
        \textbf{FlowBot} & 38\pm35 & 42\pm40 & 9\pm6 & 8\pm5 & 8\pm7 & 8\pm9 \\
        \midrule
        \textbf{\prevmethod} & \bfseries100\pm0 & \bfseries100\pm0 & \bfseries100\pm0 & \bfseries100\pm0 & \bfseries100\pm0 & \bfseries100\pm0 \\
        \textbf{Ours} & \bfseries100\pm0 & \bfseries100\pm0 & \bfseries100\pm0 & \bfseries100\pm0 & \bfseries100\pm0 & \bfseries100\pm0 \\
        \bottomrule
    \end{tabular}
\end{table}

\subsection{Evaluation Metrics}\label{sec:simulation-metrics}

To assess manipulation performance, we employ four complementary metrics that capture both effectiveness and efficiency dimensions. These metrics work together to provide a multifaceted evaluation of the manipulation task.

\paragraph*{Success rate}

As our primary measure of manipulation effectiveness, success rate employs a progress-based criterion tailored to each object type. For articulated objects with standard joints, we measure the percentage of the target displacement achieved: revolute joints are evaluated based on progress toward a \SI{60}{\degree} rotation, while prismatic joints are assessed based on progress toward a \SI{250}{\milli\meter} extension. For the Bézier curve trajectories, success is quantified as the percentage of track distance covered by the toy train from start to end.

\paragraph*{Time efficiency}

Time efficiency quantifies the temporal performance of successful manipulations, measuring the total duration from task initiation to completion. This metric holds particular significance in applications involving human-robot interaction, where swift task execution directly impacts system practicality and user experience.

\paragraph*{Action efficiency}

Action efficiency evaluates how directly each motion contributes to task progress, providing insights into the manipulation strategy's precision and economy of movement. We implement joint-specific calculations: for revolute joints, we compute normalized angular displacement relative to a \SI{60}{\degree} threshold; for prismatic joints, we measure normalized linear displacement relative to a \SI{250}{\milli\meter} threshold.

For the more complex Bézier curve trajectories, efficiency is determined by progression along the curve's parameterization \(s \in [0, 1]\), where the curve position is defined by:
\begin{equation}
    B(s) = \sum_{i=0}^{n} b_i \binom{n}{i} s^i (1-s)^{n-i},
\end{equation}
where \(b_i\) represent the control points shaping the curve, \(\binom{n}{i}\) is the binomial coefficient, and \(n\) indicates the curve order. This unified parameterization enables consistent efficiency measurement across trajectories of varying complexity.

\paragraph*{Smoothness of robot joint trajectories}

Joint trajectory smoothness serves as both an efficiency indicator and a critical operational metric in robotic manipulation. Smoother trajectories offer multiple benefits: they reflect more efficient motion control, reduce mechanical stress and wear (extending robot lifespan), enhance control precision by limiting vibrations, and improve operational safety by minimizing abrupt movements---particularly important in human-robot interaction scenarios~\cite{macfarlane2003jerk}.

To quantify smoothness while avoiding the pitfall of artificially smooth but excessively slow trajectories, we employ a time-weighted joint angle jerk. This metric measures the rate of change in joint angle acceleration scaled by task completion time, providing a balanced assessment that accounts for both smoothness and time efficiency.

Together, these four metrics provide a holistic performance profile, capturing the critical dimensions of manipulation capability from both effectiveness and efficiency perspectives.

\subsection{Simulation Results}\label{sec:simulation-results}

We begin by presenting qualitative results of \method's performance in manipulating articulated objects. Using representative examples from each category described in \cref{sec:simulation-setup}, \cref{fig:sim-results-qualitative} illustrates how the predicted handle poses effectively guide the gripper during manipulation tasks. Our observations reveal that when object motion exhibits smooth and consistent trends, the prediction model captures this behavior with high accuracy. However, as shown in the rightmost column of \cref{fig:sim-results-qualitative}, abrupt changes in object motion can occasionally introduce prediction errors, stemming from the non-causal nature of our prediction model. Nevertheless, the system demonstrates robust performance through its iterative prediction process, which continually incorporates new observations to correct these errors. These results highlight \method's practical effectiveness in handling complex manipulation scenarios commonly encountered in real-world applications. Additional qualitative results are available in the \href{\suppUrlSI}{Supplementary Video S1}.

To quantitatively assess the robustness and sensitivity of the predictive model, we evaluate the prediction residuals under varying noise levels and prediction models. While \cref{sec:methods} establishes the constant velocity model as one approach, this section extends the analysis to include constant acceleration and constant jerk models.

Real tactile sensors, such as GelSight, exhibit inherent noise on the order of micrometers. To simulate realistic conditions while assessing the robustness under greater noise, we introduce Gaussian noise with standard deviations of \SI{5.0}{\micro\meter}, \SI{50.0}{\micro\meter}, and \SI{100.0}{\micro\meter}. The prediction residual is represented by the relative handle position with respect to the robot gripper after the robot executes the predicted action. To quantify this error, we map the transformation from the Lie group to its corresponding Lie algebra and compute the 2-norm of the resulting vector.

The results are shown in \cref{fig:sim-results-prediction} with maximum prediction errors listed in \cref{tab:max_value}. All models achieve comparable performance overall. As the complexity of object kinematics increases, prediction errors increase correspondingly due to more varied motion patterns. Under low-noise conditions, higher-order models achieve better performance by capturing more complex underlying dynamics. However, they are more sensitive to noise due to the multiple differential computations involved. This tendency is clearly evident for basic prismatic and revolute joint types, but becomes less distinct in more complex kinematic scenarios. These results demonstrate that the constant-velocity model can be a robust choice, aligning with the intuition underlying previous works~\cite{davison2007monoslam,mur2015orb,scholler2020constant}. While all predictive models inevitably produce prediction residuals, \method's high-frequency operation enables rapid detection of deviations and adaptive adjustment. Rather than relying on perfect predictions, the system uses them as initial estimates within a closed-loop control framework that continuously corrects for modeling errors.

Our quantitative analysis evaluates \method using the four metrics defined in \cref{sec:simulation-metrics}.

\paragraph*{Success rate results}

As shown in \cref{tab:sim-results-success_rate}, the pre-planned method achieves considerable success rates across all tests with known kinematics. However, its performance degrades substantially under parameter estimation uncertainty, highlighting its sensitivity to modeling errors. Impedance control demonstrates effectiveness in handling moderate directional deviations without requiring known kinematics or tactile feedback. Nevertheless, its adaptive capacity becomes limited when large discrepancies exist between the planned and actual manipulation directions. The learning-based method performs well on objects with visual appearances similar to those encountered during training. However, its success rate drops to nearly zero when confronted with unfamiliar objects, revealing the generalization limitations inherent to current learning-based approaches.

Contrastingly, both tactile-informed methods, \prevmethod and \method, achieve perfect \SI{100}{\percent} manipulation success rates across all \(200\) test objects spanning six categories. This consistent performance demonstrates the fundamental effectiveness of tactile-informed approaches for articulated object manipulation. Since successful task completion is a prerequisite for meaningful efficiency evaluation, we focus our comparative analysis exclusively on methods that reliably satisfy this fundamental criterion.

\begin{figure}[t!]
    \centering
    \includegraphics[width=\linewidth]{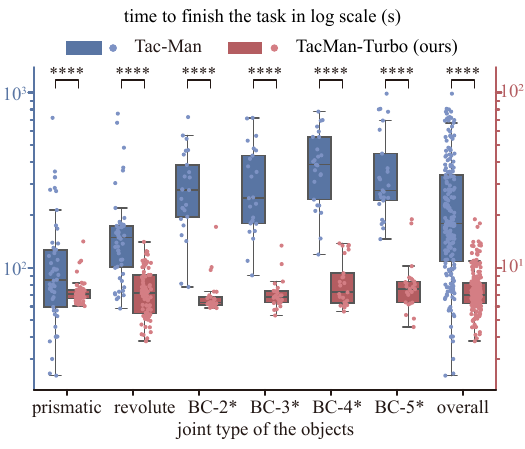}
    \caption{\textbf{Time efficiency comparison (logarithmic scale, seconds).} Evaluation of task completion times reveals \method's superior efficiency compared to \prevmethod across all tested articulation types. Our method consistently completes manipulation tasks in significantly less time, with performance advantages particularly pronounced for complex articulations. Statistical analysis confirms these improvements are highly significant (****\(p<0.0001\)) throughout all joint categories. BC-n denotes Bézier curves of order n; box plots follow conventions established in \cref{fig:sim-results-prediction}.}
    \label{fig:sim-results-time_efficiency}
\end{figure}

\begin{figure}[t!]
    \centering
    \begin{subfigure}[b]{\linewidth}
        \centering
        \includegraphics[width=\linewidth]{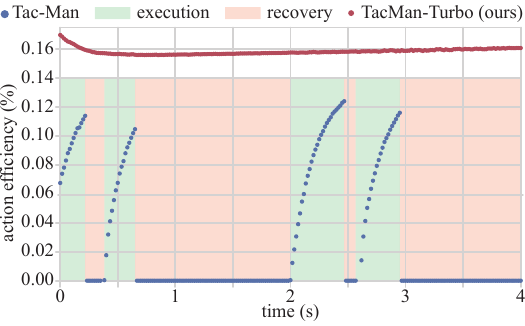}
        \caption{Temporal analysis of action efficiency during manipulation.}
        \label{fig:sim-results-action_efficiency-a}
    \end{subfigure}%
    \\%
    \begin{subfigure}[b]{\linewidth}
        \centering
        \includegraphics[width=\linewidth]{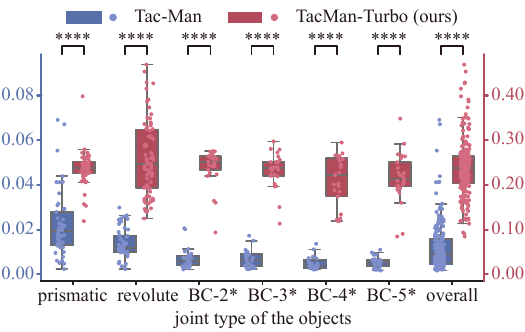}
        \caption{Mean action efficiency (\%) across object categories.}
        \label{fig:sim-results-action_efficiency-b}
    \end{subfigure}%
    \\%
    \begin{subfigure}[b]{\linewidth}
        \centering
        \includegraphics[width=\linewidth]{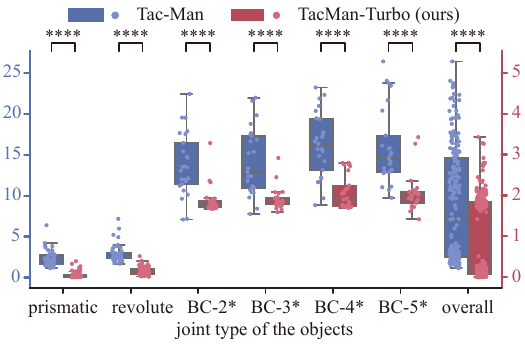}
        \caption{Relative standard deviation of action efficiency.}
        \label{fig:sim-results-action_efficiency-c}
    \end{subfigure}%
    \caption{\textbf{Analysis of action efficiency across manipulation strategies.} The results reveal \method's superior motion efficiency through three complementary measures. Temporal profiles highlight \method's continuous productive motion \vs \prevmethod's alternating execution-recovery cycles. Mean efficiency measurements confirm \method's consistently higher percentage of productive motion across all object categories. Relative standard deviation analysis demonstrates \method's more stable and predictable performance. Statistical tests across 200 objects spanning six joint categories show highly significant improvements (**** \(p<0.0001\)) in both efficiency magnitude and consistency. BC-n represents Bézier curves of order n; box plots follow conventions established in \cref{fig:sim-results-prediction}.}
    \label{fig:sim-results-action_efficiency}
\end{figure}

\begin{figure}[t!]
    \centering
    \begin{subfigure}[b]{\linewidth}
        \centering
        \includegraphics[width=\linewidth]{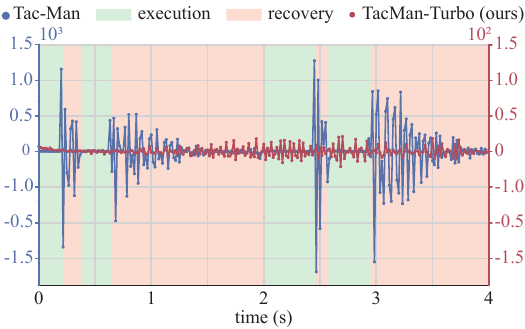}
        \caption{Time-weighted jerk profile of base joint trajectory (\si{\per\second\cubed}).}
        \label{fig:sim-results-action_jerk-a}
    \end{subfigure}%
    \\%
    \begin{subfigure}[b]{\linewidth}
        \centering
        \includegraphics[width=\linewidth]{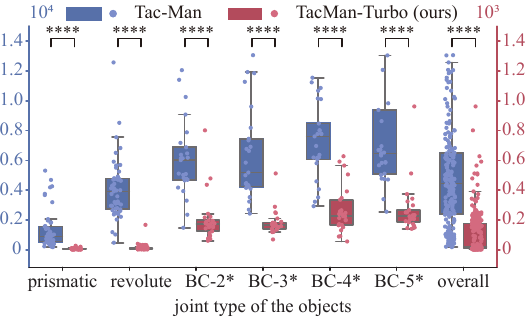}
        \caption{Mean time-weighted jerk magnitude across all categories (\si{\per\second\cubed}).}
        \label{fig:sim-results-jerk-b}
    \end{subfigure}%
    \caption{\textbf{Analysis of motion smoothness through jerk characteristics.} Motion smoothness evaluation reveals \method maintains consistently lower jerk magnitudes by eliminating the disruptive transition phases inherent to \prevmethod's approach. The temporal comparison shows \method's continuous motion profile contrasted with \prevmethod's oscillatory pattern. Statistical analysis across 200 test objects confirms \method achieves dramatic smoothness gains with standard joint types (prismatic and revolute), while improvements become more moderate with complex Bézier trajectories where challenging pose predictions occasionally necessitate corrective motions. Nevertheless, significant overall improvement (**** \(p<0.0001\)) in motion quality is observed across all categories. \textit{Note:} BC-n denotes Bézier curves of order n; box plots follow conventions established in \cref{fig:sim-results-prediction}.}
    \label{fig:sim-results-jerk}
\end{figure}

\paragraph*{Time efficiency results} 

While maintaining identical success rates, \method delivers dramatic improvements in execution speed, as detailed in \cref{fig:sim-results-time_efficiency}. Average task completion times for \method remain consistently around \SI{10}{\second} across all six object categories, compared to \prevmethod's typical requirement of over \SI{100}{\second}. Using a one-sided paired t-test on the paired differences, performance was significantly faster for \method (\(M = \SI{7.6}{\second}, SD = \SI{2.3}{\second}\)) than for \prevmethod (\(M = \SI{250.7}{\second}, SD = \SI{194.4}{\second}\)), \(t(199) = -17.7, p < 0.0001\). The one-sided \SI{95}{\percent} confidence interval for the mean paired difference was \((-\infty, -220.4]\)\,\si{\second}, and the effect size was Cohen’s \(d_\mathrm{z} = -1.3\).

\paragraph*{Action efficiency results}

The architectural differences between the two methods are most evident in their action efficiency profiles. As illustrated in \cref{fig:sim-results-action_efficiency-a}, \prevmethod's reactive approach requires dedicated recovery periods that, while necessary for correcting contact deviations, do not directly advance the manipulation task. In contrast, \method's proactive approach ensures that virtually all actions contribute directly to task progression, eliminating these non-productive recovery phases.

Our quantitative analysis examines both mean efficiency and the \acf{rsd}---the ratio of standard deviation to mean---across all trials. Results presented in \cref{fig:sim-results-action_efficiency-b,fig:sim-results-action_efficiency-c} demonstrate \method's consistently higher average efficiency with minimal variation, even when handling complex Bézier curve trajectories. Using a one-sided paired t-test on the paired differences, performance was significantly higher for \method (\(M = \SI{0.239}{\percent}, SD = \SI{0.063}{\percent}\)) than for \prevmethod (\(M = \SI{0.012}{\percent}, SD =\SI{0.011}{\percent}\)), \(t(199) = 52.6, p < 0.0001\). The one-sided \SI{95}{\percent} confidence interval for the mean paired difference was \([0.2, \infty)\)\,\si{\percent}, and the effect size was Cohen’s \(d_\mathrm{z} = 3.7\). Additionally, efficiency stability was significantly better for \method (\(M = 1.0, SD = 1.0\)) than for \prevmethod (\(M = 8.8, SD = 6.9\)), \(t(199) = -15.9, p < 0.0001\). The one-sided \SI{95}{\percent} confidence interval for the mean paired difference was \((-\infty, -7.0]\), and the effect size was Cohen’s \(d_\mathrm{z} = -1.1\).

\begin{figure}[b!]
    \centering
    \includegraphics[width=\linewidth]{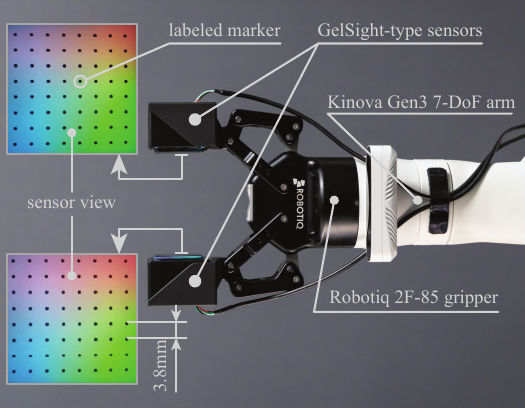}
    \caption{\textbf{Physical experimental setup.} The experimental platform consists of a Kinova Gen3 7-\ac{dof} robotic arm equipped with a Robotiq 2F-85 parallel gripper. Each gripper finger incorporates a GelSight-type tactile sensor featuring an \(8 \times 8\) marker grid for precise contact force measurement and deformation tracking.}
    \label{fig:real-exp_setup}
\end{figure}

\paragraph*{Smoothness results}

Trajectory smoothness analysis reveals fundamental behavioral differences between the two approaches. As shown in \cref{fig:sim-results-action_jerk-a}, \method demonstrates more consistent and significantly lower jerk variations compared to \prevmethod. This improvement stems directly from \method's proactive strategy, which eliminates the frequent transitions between execution and recovery stages characteristic of \prevmethod. In the reactive approach, compensatory actions to address contact deviations typically require abrupt accelerations and decelerations, resulting in higher jerk measurements.

Comprehensive testing across all \(200\) objects, summarized in \cref{fig:sim-results-jerk-b}, reveals varying degrees of improvement across object categories. \method shows the most dramatic smoothness gains with standard prismatic and revolute-joint objects, with more moderate improvements for Bézier-curve trajectories. This performance variation correlates with motion predictability---standard joint types follow consistent patterns that facilitate accurate prediction, while Bézier curves present more variable motion patterns that occasionally lead to mismatches between predicted and actual handle poses. Using a one-sided paired t-test on the paired differences, performance was significantly smoother for \method (\(M = \SI{109.8}{\per\second\cubed}, SD = \SI{147.7}{\per\second\cubed}\)) than for \prevmethod (\(M = \SI{4291.4}{\per\second\cubed}, SD = \SI{3722.49}{\per\second\cubed}\)), \(t(199) = -18.3, p < 0.0001\). The one-sided \SI{95}{\percent} confidence interval for the mean paired difference was \((-\infty, -3803.4]\)\,\si{\per\second\cubed}, and the effect size was Cohen’s \(d_\mathrm{z} = -1.3\).

\section{Real-World Experiments}\label{sec:real-world_exp}

While simulation studies establish \method's theoretical advantages, real-world deployment introduces critical challenges absent in simulation: sensor noise, execution uncertainty, and physical constraints. These factors can significantly impact control performance, necessitating comprehensive physical validation. This section presents our real-world experimental evaluation in four complementary parts: experimental setup (\cref{sec:real-exp_setup}), testing protocols (\cref{sec:real-exp_protocol}), performance analysis (\cref{sec:real-exp_results}), and validation on everyday objects (\cref{sec:additional_results})---including categories not represented in standard articulated object datasets.

\begin{figure}[t!]
    \centering
    \begin{subfigure}[b]{.33\linewidth}
        \centering
        \includegraphics[height=2.9cm]{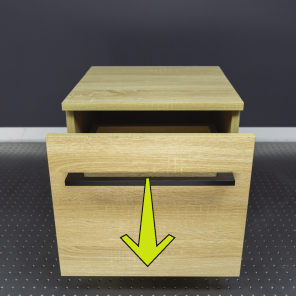}
        \caption{}
        \label{fig:real-objects-a}
    \end{subfigure}%
    \hfill%
    \begin{subfigure}[b]{.33\linewidth}
        \centering
        \includegraphics[height=2.9cm]{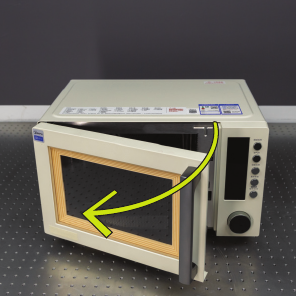}
        \caption{}
        \label{fig:real-objects-b}
    \end{subfigure}%
    \hfill%
    \begin{subfigure}[b]{.33\linewidth}
        \centering
        \includegraphics[height=2.9cm]{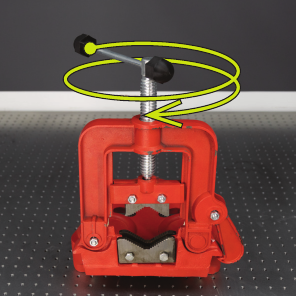}
        \caption{}
        \label{fig:real-objects-c}
    \end{subfigure}%
    \caption{\textbf{Test objects for real-world experiments.} The experimental validation employs three articulated objects with distinct kinematic characteristics: (a) a wooden drawer representing pure prismatic joint motion along a linear path, (b) a microwave oven featuring revolute joint dynamics through its door hinge mechanism, and (c) a bench vise with a hand crank generating complex helical motion. These objects collectively challenge the manipulation system across fundamental mechanical joint categories, from simple linear translation to complex compound movements requiring sophisticated trajectory prediction.}
    \label{fig:real-objects}
\end{figure}

\subsection{Real-World Experimental Setup}\label{sec:real-exp_setup}

Our physical validation framework integrates three essential components designed to evaluate \method's performance under realistic conditions: a robotic manipulation platform, high-fidelity tactile sensing, and diverse test objects with varied kinematic properties.

\paragraph*{Robot system setup}

Manipulating articulated objects with varied handle paths requires full 6-\ac{dof} movement capability in Cartesian space. Our platform employs a 7-\ac{dof} Kinova Gen3 robotic arm paired with a Robotiq 2F-85 gripper, providing the necessary workspace coverage and manipulation flexibility. To enable precise contact sensing---crucial for our tactile-based approach---we modified the gripper by replacing its standard pads with two GelSight-type tactile sensors. This integrated system, illustrated in \cref{fig:real-exp_setup}, combines motion versatility with tactile feedback. The tactile sensor’s pose relative to the robot wrist is determined from the robot’s mechanical model. The internal camera of the GelSight-type sensor is then calibrated using OpenCV’s built-in routines: we estimate the camera intrinsics and lens distortion coefficients by imaging laser-cut black markers that serve as calibration targets.

Overall, the tactile sensor updates contact information at \SI{120}{\hertz}, while the offset velocity is computed at \SI{50}{\hertz} and transmitted to Kinova's low-level controller, which operates at \SI{1000}{\hertz}. All computations were performed on Ubuntu 22.04 with an AMD Ryzen 9 5950X CPU and an NVIDIA GeForce RTX 3090 GPU. Over \num{10000} test steps, the computation time from acquiring a new tactile image to producing the velocity command was \SI{3.70}{\milli\second}\(\pm\)\SI{0.85}{\milli\second}. During these tests, the CPU load was \(\SI{13.36}{\percent}\pm\SI{3.39}{\percent}\), and the GPU load was \(\SI{56.75}{\percent}\pm\SI{2.04}{\percent}\). This low latency enables the controller to receive and react to new feedback in real time. Further discussion of the update frequency is provided in \cref{sec:discussion-base_velocity}.

\begin{figure*}[t!]
    \centering
    \includegraphics[width=\linewidth]{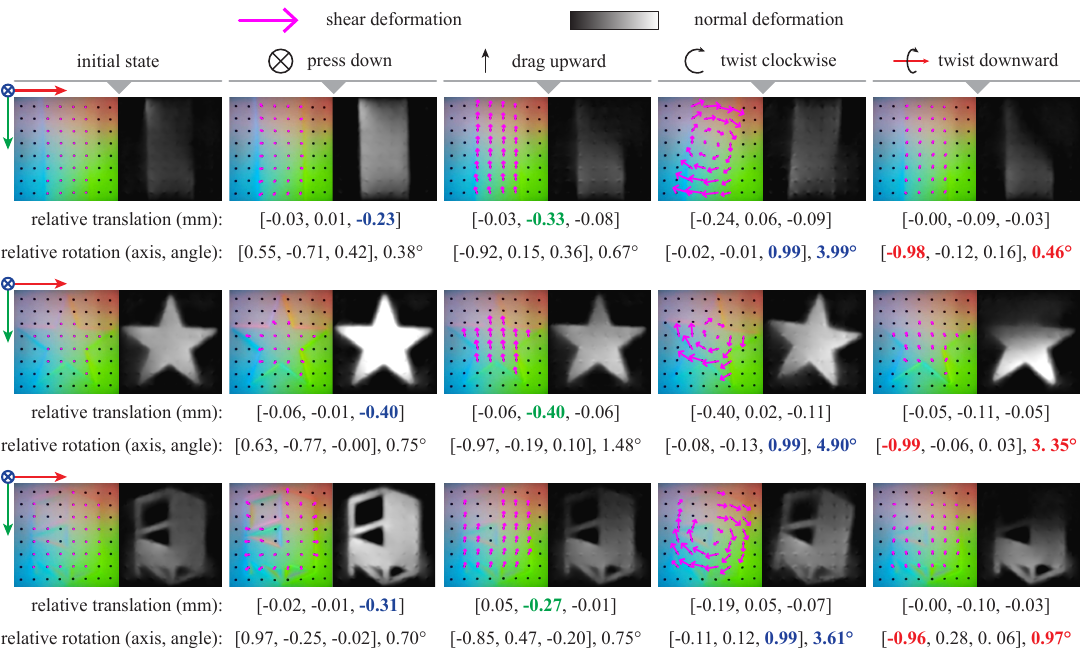}
    \caption{\textbf{Tactile-informed pose estimation of in-hand objects.} Precise tracking of object movements during manipulation is demonstrated through tactile feedback across three different objects subjected to five distinct manipulations: initial state, pressing down, dragging upward, and two types of twisting motions. Contact points (marked with \textcolor[RGB]{255,0,255}{purple dots}) are tracked to compute relative pose transformations using \cref{eq:transformation}. The colored grids display marker displacement patterns while the grayscale images reveal normal force distribution. Quantitative relative translations (mm) and rotations (axis-angle representation) below each image pair confirm accurate capture of subtle pose changes. Visualization arrows (scaled 7× for clarity) illustrate movement direction and magnitude, showing consistent correspondence between applied manipulations and detected transformations across varied object geometries.}
    \label{fig:real-pose_estimation}
\end{figure*}

\begin{figure*}[t!]
    \centering
    \includegraphics[width=0.99\linewidth]{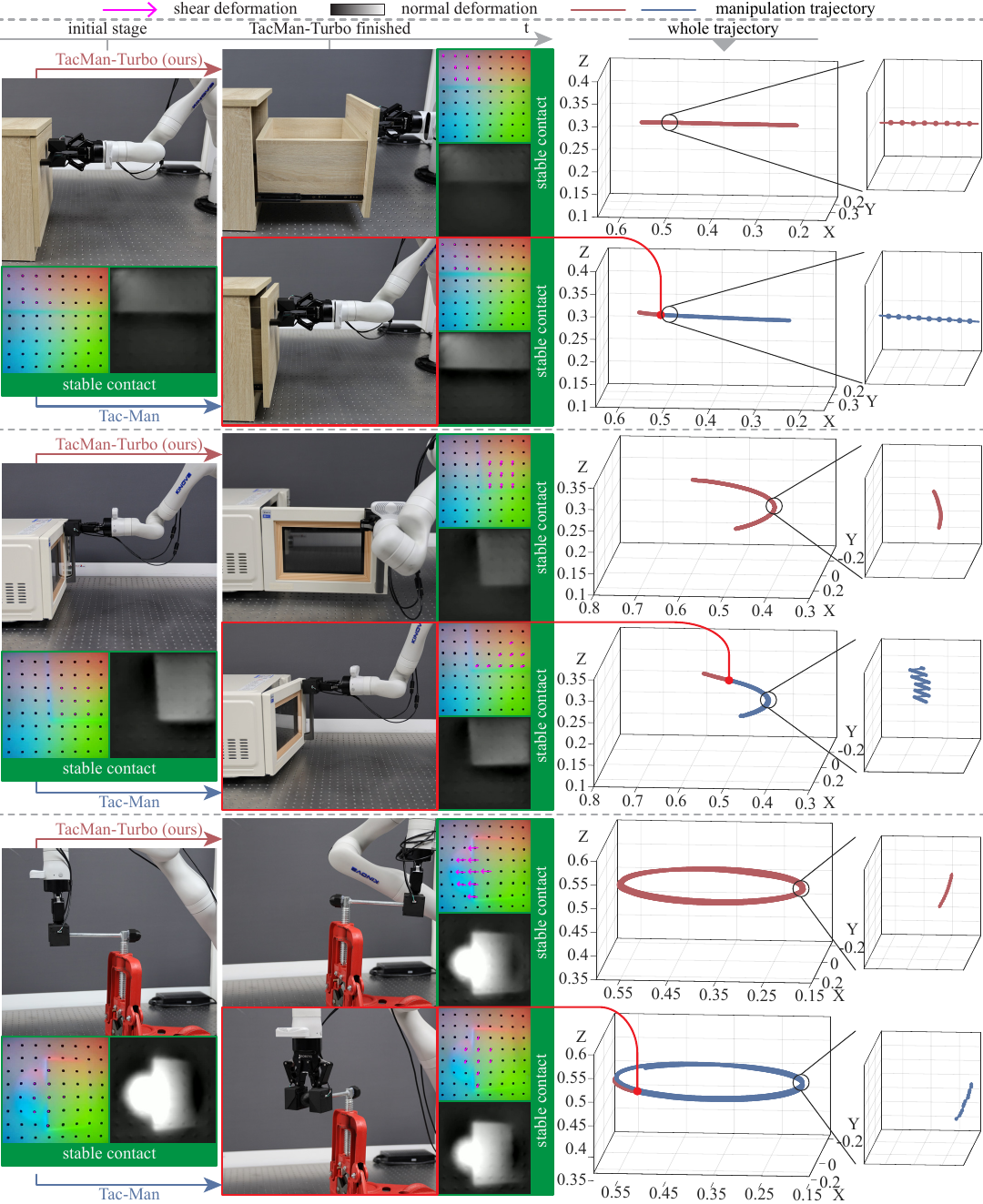}
    \caption{\textbf{Real-world experiment results.} Comparative analysis across three articulated objects (drawer, microwave, vise) demonstrates successful manipulation by both methods with clear performance differences. Each row presents a complete manipulation sequence showing initial stable contact (left), manipulation stages (middle), and corresponding 3D trajectories (right). Tactile feedback visualizations display contact force distributions through marker displacement patterns (colored grids) and normal force intensity (grayscale images). While both approaches successfully complete all tasks, \method consistently delivers more direct trajectories, resulting in reduced completion time and smoother motion profiles. Detailed comparisons are available in the \href{\suppUrlSIII}{Supplementary Video S3}.}
    \label{fig:real-results}
\end{figure*}

\begin{figure}[t!]
    \centering
    \begin{subfigure}[b]{.24\linewidth}
        \centering
        \includegraphics[width=\linewidth]{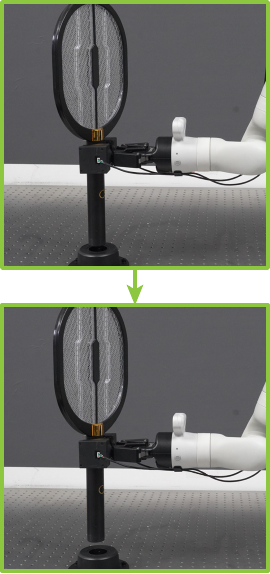}
        \caption{}
        \label{fig:additional_results-a}
    \end{subfigure}%
    \hfill%
    \begin{subfigure}[b]{.24\linewidth}
        \centering
        \includegraphics[width=\linewidth]{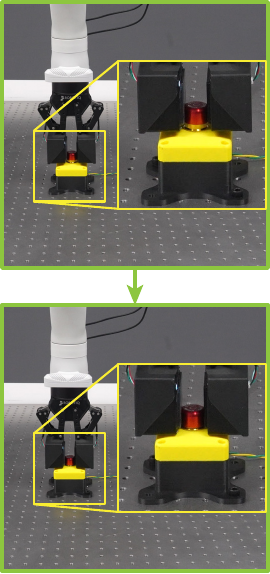}
        \caption{}
        \label{fig:additional_results-b}
    \end{subfigure}%
    \hfill%
    \begin{subfigure}[b]{.24\linewidth}
        \centering
        \includegraphics[width=\linewidth]{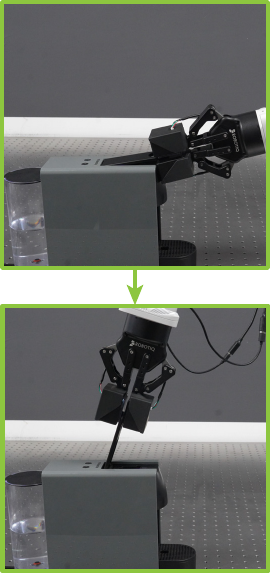}
        \caption{}
        \label{fig:additional_results-c}
    \end{subfigure}%
    \hfill%
    \begin{subfigure}[b]{.24\linewidth}
        \centering
        \includegraphics[width=\linewidth]{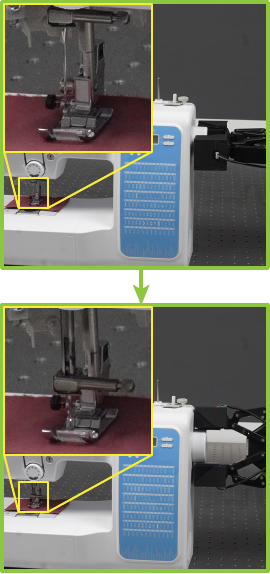}
        \caption{}
        \label{fig:additional_results-d}
    \end{subfigure}%
    \caption{\textbf{Tests on various common household objects.} Our \method approach successfully generalizes to diverse everyday objects without requiring additional tuning. The system effectively manipulates: (a) a battery charging dock requiring precise device extraction, (b) a power switch with discrete on/off state transitions, (c) a coffee maker with a rotational lid mechanism, and (d) a sewing machine featuring complex mechanical linkages. Before (top) and after (bottom) images demonstrate successful manipulation for each object. These tests validate \method's ability to adapt to varied kinematic constraints and interaction requirements found in common household items, confirming its practical applicability beyond the primary test objects. Detailed manipulation processes are available in the \href{\suppUrlSIV}{Supplementary Video S4}.}
    \label{fig:additional_results}
\end{figure}

\paragraph*{Tactile sensors}\label{sec:real-tactile_sensors}

Accurate contact information between the object and robot is essential for \method's operation. Our implementation utilizes two GelSight-type tactile sensors~\cite{yuan2017gelsight}, each featuring an \(8\times8\) grid of black markers (\cref{fig:real-exp_setup}). These markers enable precise computation of the relative transformation matrix between gripper and handle through point correspondence matching, as formalized in \cref{eq:transformation}. 

To validate the sensing system's reliability, we conducted controlled perturbation tests with in-hand objects (\cref{fig:real-pose_estimation}). The computed transformations consistently aligned with applied perturbation directions, confirming the system's ability to accurately estimate relative pose changes---a capability fundamental to the prediction framework described in \cref{sec:method-pose_prediction}.

Selecting contact points requires careful parameter tuning to ensure accurate identification within the contact region; see \cref{eq:contact_points}. As noted by Zhao~\etal~\cite{zhao2024tac}, the transition zone between contact and non-contact regions can exhibit substantial normal deformation due to the soft sensor’s compliance, yet it does not reliably reflect changes in handle position. This makes a fixed threshold \(\epsilon\) ineffective.

To mitigate this, we adapt \(\epsilon\) in \cref{eq:contact_points} during each trial, choosing the smallest value that yields exactly nine contact points in the initial contact set \(\mathcal{C}_i\). Given that our setup places at most eight markers on any single line, this choice satisfies the constraints in \cref{eq:num_contact_points,eq:noncollinear_constraint} by preventing collinearity, thereby enabling robust computation of contact deviations.

\paragraph*{Test articulated objects}

Following our simulation methodology, we evaluate \method across three kinematically distinct categories: a drawer with prismatic joints (\cref{fig:real-objects-a}), a microwave oven with revolute joints (\cref{fig:real-objects-b}), and a vise featuring a helical handle path (\cref{fig:real-objects-c}). This selection enables systematic assessment across fundamental joint types while incorporating real-world mechanical complexities such as friction variations and mechanical play.

\subsection{Real-World Experiment Protocol}\label{sec:real-exp_protocol}

A fair comparison between manipulation approaches requires standardized testing conditions that isolate algorithmic differences from experimental variations. We implemented a rigorous protocol for evaluating \method against \prevmethod across all real-world scenarios. 

Each experimental trial began with identical initialization conditions: manually established stable grasps on object handles (illustrated in the left column of \cref{fig:real-results}), ensuring consistent starting contact states. We maintained controlled experimental parameters by testing both methods on the same physical objects with identical initial interaction directions, eliminating potential confounding variables. For the grasping force, since a larger force ensures the non-slipping assumption, we used the lesser of two values: the gripper's maximum force or the tactile sensor's force tolerance. In the case of the selected Robotiq 2F-85 gripper, its maximum force is significantly lower than the tactile sensor's capacity. Therefore, we consistently applied the gripper's maximum force.

Throughout each manipulation sequence, we recorded two complementary performance metrics: task completion time (measuring overall efficiency) and complete gripper trajectories (capturing motion quality). This dual measurement approach allows a comprehensive assessment of both the temporal efficiency and spatial characteristics of each manipulation strategy.

\subsection{Real-World Experiment Results}\label{sec:real-exp_results}

Our experimental results, presented in \cref{fig:real-results}, provide evidence of both methods' manipulation capabilities while revealing significant performance distinctions between their approaches.

\paragraph*{Manipulation effectiveness}

Both \method and \prevmethod successfully manipulated all test objects shown in \cref{fig:real-objects}, confirming the fundamental effectiveness of tactile-based strategies for articulated object manipulation. Each method maintained stable contact throughout the manipulation process and completed all assigned tasks. The resulting motion trajectories (middle-right column, \cref{fig:real-results}) closely followed expected kinematic patterns: linear motion for the prismatic-joint drawer, circular arcs for the revolute-joint microwave door, and extended helical paths for the vise handle. Minor deviations primarily stemmed from manufacturing tolerances inherent in real-world objects.

\paragraph*{Performance differences}

Despite similar success rates, time efficiency analysis revealed dramatic performance differences between the two approaches. As captured in the middle-left column of \cref{fig:real-results}, \method completed manipulation tasks substantially faster, while \prevmethod had barely initiated object movement at the same point in time. This efficiency gap stemmed from fundamental differences in operational strategies evident in detailed trajectory analysis (right column, \cref{fig:real-results}).

The operational differences were most pronounced in objects requiring frequent directional adjustments. For the drawer's linear motion, where the initial interaction direction naturally aligned with the object's inherent trajectory, both methods demonstrated comparable path following. However, for revolute and helical motions requiring continuous directional adaptation, the methods exhibited markedly different behaviors:

\prevmethod employed a stepwise exploratory approach, advancing until a predefined threshold was reached. During this process, motion direction gradually drifted from the optimal trajectory, resulting in reduced action efficiency. While this drift was eventually corrected during recovery phases, the repeated cycles of exploration and correction produced non-smooth, zig-zag motion patterns with frequent velocity reversals.

In contrast, \method leveraged predictive modeling to proactively adjust its motion direction. This continuous adaptation achieved consistently high action efficiency, smoother trajectories, and significantly faster task completion by eliminating the need for separate recovery phases.

These differences in execution time, action efficiency, and trajectory smoothness highlight \method's superior performance, successfully translating its theoretical advantages from simulation to real-world application. The results demonstrate \method's robustness in practical scenarios, effectively addressing challenges including sensor noise, execution uncertainty, and physical constraints. For detailed visualization of the manipulation process, including tactile sensor feedback throughout execution, we direct readers to the \href{\suppUrlSII}{Supplementary Video S2} and \href{\suppUrlSIII}{Supplementary Video S3}.

\subsection{Additional Tests on Everyday Objects}\label{sec:additional_results}

To assess \method's generalization capabilities beyond laboratory test objects, we extended our evaluation to include diverse household items encountered in daily environments. These objects, all amenable to secure manipulation using a parallel gripper, present varied kinematic structures and functional challenges not represented in standard datasets.

A battery charging dock (\cref{fig:additional_results-a}) exemplifies objects with linear motion paths, where precise trajectory following is essential for secure device removal and to prevent potential damage. A power switch (\cref{fig:additional_results-b}) represents safety-critical mechanisms with discrete state transitions, requiring reliable manipulation despite high resistance forces. A coffee maker (\cref{fig:additional_results-c}) involves rotational motion, demanding adaptive handling to ensure smooth operation. Lastly, a sewing machine (\cref{fig:additional_results-d}) introduces intricate motion patterns with varying resistance caused by its mechanical linkages and operational states.

These objects are not fully represented in standard articulated object datasets, highlighting \method's ability to generalize to novel, real-world mechanisms. Starting from stable initial grasps, \method enabled the robot to successfully execute practical tasks: safely removing devices from charging docks, reliably operating emergency power switches, preparing coffee makers for use, and manipulating sewing machines for basic textile operations.

The consistent success across this diverse set of previously unseen objects demonstrates \method's robustness and effectiveness in real-world applications. Its ability to handle varying physical properties, motion patterns, and functional requirements underscores its practical utility for everyday manipulation tasks. For detailed visualization of the manipulation process, including tactile sensor feedback throughout execution, we direct readers to the \href{\suppUrlSIV}{Supplementary Video S4}.

\begin{figure*}[t!]
    \centering
    \includegraphics[width=\linewidth]{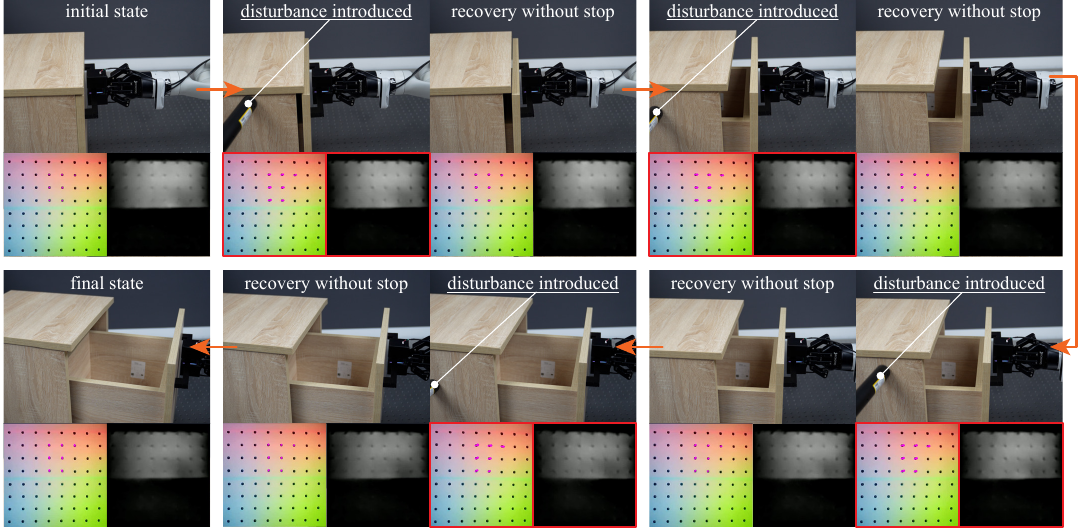}
    \caption{\textbf{Manipulation under disturbance.} The efficiency gains enabled by \method do not compromise its robustness against unexpected disturbances that frequently occur in human-centric environments. Unlike previous reactive paradigms~\cite{zhao2024tac} that require stopping for reactive adjustments, \method can maintain manipulation progress continuously. Detailed manipulation processes are available in the \href{\suppUrlSV}{Supplementary Video S5}.}
    \label{fig:with_disturbance}
\end{figure*}

\section{Discussions}\label{sec:discussion}

\subsection{Comparison with \prevmethod}

Both \method and \prevmethod leverage robot-object contact as the state representation. This approach ensures robustness without requiring pre-defined object kinematics for articulated object manipulation. The effectiveness of this strategy is comprehensively demonstrated through simulation results (\cref{tab:sim-results-success_rate}) and validated in real-world experiments (\cref{fig:real-results}).

In the tactile control context, this modeling framework facilitates direct generation of robot Cartesian motions, eliminating the need for parameter tuning such as PID gain selection. Furthermore, unlike other approaches that also directly generate robot Cartesian motions, this framework avoids reliance on manually-defined tactile features. Instead, it leverages contact points for frame registration, making it inherently robust and agnostic to variations in handle shapes and structures.

Although both approaches share a similar state representation, they address the manipulation problem fundamentally differently. \prevmethod adopts a \textbf{reactive} strategy, prioritizing compensation for the current contact deviation---denoted as the deviation at step \(i\). This approach temporarily halts object manipulation and focuses exclusively on deviation correction. Consequently, the process naturally divides into two distinct stages: execution and recovery, as evident in \cref{fig:sim-results-action_efficiency-a} and the zig-zag trajectories observed in \cref{fig:real-results}. While this formulation enhances robustness by ensuring timely compensation, it introduces inefficiencies, as the recovery stage does not directly advance the manipulation task. Additionally, transitions between execution and recovery stages produce non-smooth trajectories, as highlighted in \cref{fig:sim-results-action_jerk-a} and \cref{fig:real-results}. Although \prevmethod incorporates multi-step exploration in real deployments to mitigate these limitations, the fundamental drawbacks of the reactive formulation remain.

In contrast, \method's core motivation explores whether robust manipulation can be achieved without interrupting the object interaction process. This question drives the development of \method's \textbf{proactive} formulation. Unlike reactive approaches, this formulation must account not only for the current \(i\)-step contact deviation but also anticipate the impact of the future \(i+1\)-step contact deviation caused by continuous, non-halting manipulation, as described in \cref{eq:formulation}.

This non-causal formulation requires estimating future object states---a challenge that reactive approaches cannot address, as they treat contact deviations solely as instantaneous spatial error signals. \method overcomes this through a temporal perspective: by analyzing sequential contact observations as time-series data, we extract object local kinematic information inherently embedded within tactile signals. This temporal interpretation enables \method to predict future object states without requiring pre-specified kinematic models, allowing proactive velocity modulation that minimizes predicted future deviations.

This proactive formulation streamlines the previous execute-and-recover cycle into a continuous control loop. As a result, it delivers significant performance gains, including an order-of-magnitude improvement in time efficiency, consistently high action efficiency, and smooth trajectories free from acceleration-deceleration cycles---all while maintaining manipulation effectiveness. Furthermore, this framework maintains robustness against unexpected disturbances that frequently occur in human-centric environments while continuing the manipulation process without stopping for reactive compensation, as shown in \cref{fig:with_disturbance} and \href{\suppUrlSV}{Supplementary Video S5}.

\begin{figure*}[t!]
    \centering
     \begin{subfigure}[b]{.5\linewidth}
        \centering
        \includegraphics[width=\linewidth]{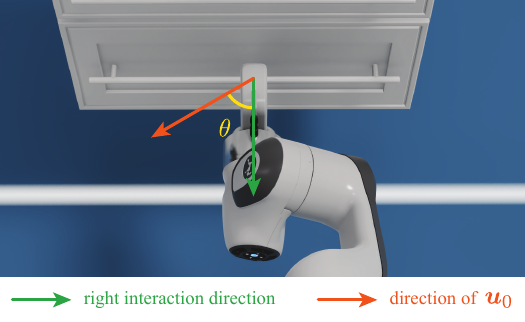}
        \caption{Simulation setup.}
        \label{fig:discussion-v_direction-a}
    \end{subfigure}%
    \begin{subfigure}[b]{.5\linewidth}
        \centering
        \includegraphics[width=\linewidth]{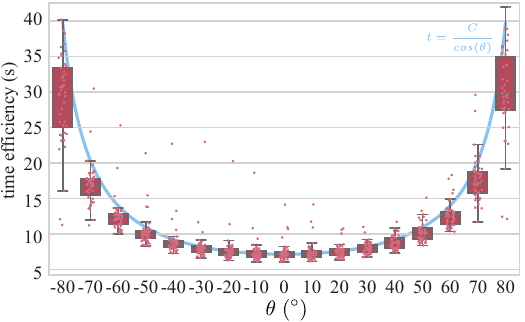}
        \caption{Time efficiency varying with \(\theta\).}
        \label{fig:discussion-v_direction-b}
    \end{subfigure}%
    \\%
    \begin{subfigure}[b]{.5\linewidth}
        \centering
        \includegraphics[width=\linewidth]{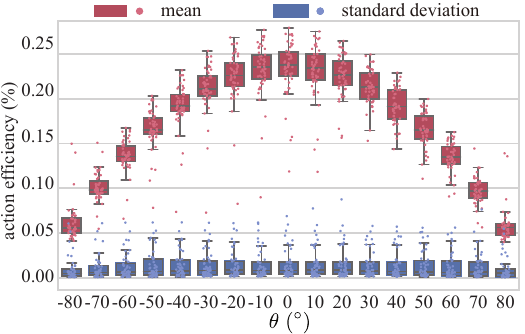}
        \caption{Action efficiency varying with \(\theta\).}
        \label{fig:discussion-v_direction-c}
    \end{subfigure}%
    \begin{subfigure}[b]{.5\linewidth}
        \centering
        \includegraphics[width=\linewidth]{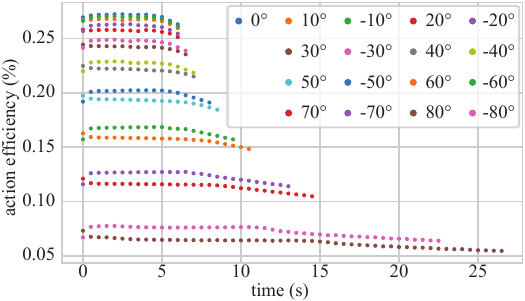}
        \caption{An example of temporal action efficiency variation with \(\theta\).}
        \label{fig:discussion-v_direction-d}
    \end{subfigure}%
    \caption{\textbf{Effects of base velocity direction.} We investigate how initial directional alignment impacts manipulation performance. (a) The simulation setup measures the angle \(\theta\) between the correct interaction direction (green arrow) and the initial motion direction \(\boldsymbol{u}_0^{g_i}\) (orange arrow). (b) Time efficiency results reveal significant performance degradation as \(\theta\) increases in magnitude, with optimal performance near \(\theta=0^{\circ}\). The degradation closely follows a $1/\cos{(\theta)}$ trend. The constant $C$ in the figure denotes the median time efficiency at $\theta=0^{\circ}$. (c) Action efficiency peaks when the initial direction closely aligns with the correct interaction direction and gradually decreases with increasing angular deviation. (d) The temporal progression further illustrates how different initial angles affect performance throughout the manipulation process, with properly aligned initial directions maintaining consistently higher efficiency. These findings highlight the critical importance of directional alignment in manipulation tasks. Box plots follow conventions established in \cref{fig:sim-results-prediction}.}
\end{figure*}

\subsection{Base Velocity Effects and Acquisition}\label{sec:discussion-base_velocity}

The performance of \method depends critically on the initial base velocity \(\boldsymbol{u}_0^{g_i}\) introduced in \cref{sec:method-formulation}. While our primary experiments utilized manually assigned values for this parameter, understanding both its effects on system performance and methods for its automated acquisition has significant implications for practical deployment. We analyze the magnitude and direction components separately to provide a more structured understanding of their respective influences.

\paragraph*{Magnitude effect and acquisition} 

Velocity magnitude directly impacts manipulation efficiency, with higher values generally enabling faster task completion. However, two fundamental constraints limit the applicable magnitude: the manipulator's physical capabilities and the tactile sensors' characteristics.

The manipulator constraint is straightforward---\(\boldsymbol{u}_0^{g_i}\) cannot exceed the maximum permissible end-effector velocity. The tactile sensor constraint presents more nuanced considerations. For accurate contact deviation detection, differences in consecutive sensor readings (induced by \(\boldsymbol{u}_0^{g_i}\)) must remain below the threshold that would trigger relative slippage between sensor and object. The most limiting scenario occurs when \(\boldsymbol{u}_0^{g_i}\) is perpendicular to the optimal interaction direction, as this configuration contributes maximally to contact deviation. Under these conditions, the constraint is formalized as:
\begin{equation}
    \boldsymbol{u}_0^{g_i} \leq \boldsymbol{\gamma} f,
\end{equation}
where \(\boldsymbol{\gamma}\) represents the critical slippage threshold and \(f\) denotes sampling frequency. This formulation reveals two pathways to enabling higher base velocities and thus improving manipulation efficiency: increasing surface roughness (which raises \(\boldsymbol{\gamma}\)) or elevating the sampling frequency.

Our implementation employed vision-based tactile sensors limited by their image sensors to approximately \SI{120}{\hertz} sampling frequency. Alternative sensing technologies, such as capacitive or 6D force/torque sensors, can operate at frequencies exceeding \SI{1000}{\hertz}, suggesting promising opportunities for substantial efficiency gains through \method adaptation to these advanced modalities.

\paragraph*{Direction effect and acquisition}

The direction of base velocity profoundly influences manipulation efficiency. Greater alignment between \(\boldsymbol{u}_0^{g_i}\) and the object's optimal interaction direction yields higher first-step action efficiency (as defined in \cref{sec:simulation-setup}). This initial efficiency advantage propagates through subsequent steps, potentially maintaining elevated performance throughout the entire manipulation process.

To quantify this relationship, we conducted systematic simulations across 50 prismatic-joint objects (\cref{fig:discussion-v_direction-a}), varying the angle \(\theta\) between \(\boldsymbol{u}_0^{g_i}\) and the optimal direction. The results confirm our theoretical expectations: both time efficiency (\cref{fig:discussion-v_direction-b}) and action efficiency (\cref{fig:discussion-v_direction-c}) reach maximum values at \(\theta=0^{\circ}\) and deteriorate as \(\theta\) increases. The time efficiency degradation closely follows $1/\cos{(\theta)}$ trend, consistent with the scaling of the correct velocity component’s magnitude. This implies that as \(\theta\) approaches $90^{\circ}$, the time cost grows unboundedly---the manipulator maintains contact but ceases to produce meaningful object displacement. Furthermore, \cref{fig:discussion-v_direction-d} reveals that per-step action efficiency maintains remarkable consistency throughout the manipulation process, closely tracking the first-step efficiency and achieving its peak at perfect alignment (\(\theta=0\)).

These findings not only validate our theoretical framework but also highlight the critical importance of appropriate base velocity direction selection. They further suggest promising research directions for automating preliminary directional information acquisition. Recent advances in vision-based direction estimation~\cite{eisner2022flowbot3d,yu2024gamma,wang2024rpmart} offer particularly promising pathways. While such methods inevitably introduce some estimation noise, they enhance \method's operational autonomy. Moreover, our framework's inherent robustness could effectively compensate for moderate directional inaccuracies, as demonstrated by the gradual performance degradation shown in \cref{fig:discussion-v_direction-b}.

\section{Conclusion and Future Work}\label{sec:conclusion}

This paper introduces \method, a novel proactive control framework that successfully mitigates the persistent trade-off between efficiency and effectiveness in articulated object manipulation. By fundamentally reframing contact deviations as valuable kinematic information rather than mere error signals, our approach enables proactive control adjustments that significantly enhance manipulation efficiency while maintaining robust performance across diverse articulated mechanisms.

Our evaluation, including varied kinematic structures and multiple real-world experiments, demonstrates that \method achieves perfect manipulation success rates (\SI{100}{\percent}) while delivering substantial improvements in time efficiency, action efficiency, and trajectory smoothness compared to previous approaches. The statistical significance of these performance gains (p-values \(< 0.0001\)) confirms that \method effectively mitigates the effectiveness-efficiency trade-off that has challenged existing manipulation methods.

This work represents an important advancement toward practical robot deployment in human-centric environments, where robots must manipulate articulated objects reliably and efficiently without precise pre-provided kinematic models. The ability to extract and leverage kinematic insights directly from tactile feedback enables robots to adapt to uncertain object structures while executing smooth, efficient movements---a capability essential for robots operating effectively in dynamic, everyday settings alongside humans.

Several promising directions for future research emerge from this work. First, developing adaptive methods for base velocity optimization could further enhance efficiency by dynamically adjusting control parameters in response to object characteristics. Second, extending the framework to support diverse tactile sensing modalities would broaden its applicability across robot platforms with different sensing capabilities. Finally, the current research primarily focuses on rigid, single-joint articulated objects. Third, while our current focus is on rigid, single-joint articulated objects, effectively handling soft, multi-part articulated systems in cluttered environments remains an important and promising challenge. These enhancements will not only further boost the framework's efficiency but also facilitate its application in increasingly complex, unstructured real-world environments.

\section*{Acknowledgment}

Our sincere thanks go to Yuanhong Zeng (UCLA) and Qian Long (UCLA) for their suggestion in driving Kinova, Lei Yan (LeapZenith AI Research) for mechanical design support, Yida Niu (PKU) for demo shooting, Saiyao Zhang (UCAS) for simulation support, and Ms. Hailu Yang (PKU) for her assistance in procuring every piece of raw material necessary for this research.

\bibliographystyle{ieeetr}
\bibliography{reference_header,reference}

\end{document}

%% file: config.tex
\usepackage{dblfloatfix}
\crefname{ineq}{Ineq.}{Ineqs.}
\creflabelformat{ineq}{#2{\upshape(#1)}#3} 

\newcommand{\method}{TacMan-Turbo\xspace}
\newcommand{\prevmethod}{Tac-Man\xspace}
\newcommand{\transpose}{\mathsf{T}}

\acrodef{dof}[DoF]{degree of freedom}
\acrodefplural{dof}[DoFs]{degrees of freedom}
\acrodef{iqr}[IQR]{Interquartile Range}
\acrodef{rsd}[RSD]{Relative Standard Deviation}
\acrodef{mpc}[MPC]{Model Predictive Control}
\acrodef{tpc}[TPC]{Tactile Predictive Control}
\acrodef{ptc}[PTC]{Proactive Tactile Control}

\usepackage{orcidlink}
\usepackage[detect-weight,mode=text]{siunitx}
\newcommand{\suppUrlSI}{https://player.vimeo.com/video/1106998274?h=1c83e4d580}
\newcommand{\suppUrlSII}{https://player.vimeo.com/video/1106998395?h=1b17dc99f5}
\newcommand{\suppUrlSIII}{https://player.vimeo.com/video/1106998713?h=e4df8b2a99}
\newcommand{\suppUrlSIV}{https://player.vimeo.com/video/1106999100?h=0b4635e4f9}
\newcommand{\suppUrlSV}{https://player.vimeo.com/video/1106999231?h=6378522bcd}